\documentclass{article}

\PassOptionsToPackage{numbers, compress}{natbib}

\usepackage[preprint]{neurips_2026}

\usepackage[utf8]{inputenc} % allow utf-8 input
\usepackage[T1]{fontenc}    % use 8-bit T1 fonts
\usepackage{hyperref}       % hyperlinks
\usepackage{url}            % simple URL typesetting
\usepackage{booktabs}       % professional-quality tables
\usepackage{amsfonts}       % blackboard math symbols
\usepackage{nicefrac}       % compact symbols for 1/2, etc.
\usepackage{microtype}      % microtypography
\usepackage{xcolor}         % colors
\usepackage{graphicx}
\usepackage{tikz}
\usepackage{multirow}
\usepackage{placeins}
\usepackage{amsmath}
\usepackage{wrapfig}
\usepackage{caption}
\usepackage{subcaption}
\usepackage{cleveref}
\usepackage[most]{tcolorbox}

\newtcolorbox{insight}[1][]{
  colback=blue!5,
  colframe=blue!50!black,
  boxrule=0.5pt,
  arc=2pt,
  left=6pt, right=6pt, top=4pt, bottom=4pt,
  fonttitle=\bfseries,
  #1
}

\AddToHook{cmd/appendix/before}{\crefalias{section}{appendix}}
\crefname{appendix}{Appendix}{Appendices}
\Crefname{appendix}{Appendix}{Appendices}

\usetikzlibrary{positioning, arrows.meta, fit, calc, backgrounds, shapes.geometric, fadings, decorations.pathreplacing, decorations.markings, shapes.arrows, patterns}

\title{Attention Drift: What Autoregressive Speculative Decoding Models Learn \thanks{Code: \url{https://github.com/Dogacel/Attention-Drift}}}

\author{%
  Doğaç Eldenk \\
  Northwestern University \\
  \And
  Payal Mohapatra \\
  Northwestern University \\
  \And
  Yigitcan Comlek \\
  GE Aerospace \\
  \And
  Kaan Oktay \\
  fal \\
  \And
  Hongyang Zhang \\
  University of Waterloo \\
  \And
  Stephen Xia \\
  Northwestern University
}

\begin{document}

\maketitle

\begin{abstract}
    Speculative decoding accelerates LLM inference by drafting future tokens with a small model. However, drafter models degrade sharply under template perturbation and long-context inputs. We identify a previously-unreported phenomenon we call \textbf{attention drift}: as the drafter generates successive tokens within a speculation chain, attention progressively moves from the prompt onto its own recently-generated tokens. This phenomenon is observed across both \emph{EAGLE3} drafters and \emph{MTP heads}, suggesting drift is a property of drafter designs. We trace this to the unnormalized residual path between chain steps: the drafter's hidden state magnitude grows monotonically with chain depth, which exhibits dynamics consistent with additional pre-norm transformer layers stacked on the target rather than as a standalone autoregressive predictor. In order to limit the growth, we propose two architectural changes: Post-norm on the drafter hidden states and per-hidden-state RMSNorm after capturing target hidden states. Our interventions improve acceptance length over the current leading model, pre-norm EAGLE3, by up to $2\times$ under template perturbation, $1.18\times$ on long-context tasks, and $1.10\times$ on seven standard benchmarks spanning multi-turn chat, math, and coding. Our changes also allow shorter train-time-test depths to generalize over longer drafting sequences.
\end{abstract}

\begin{figure}[h]
    \centering
    \begin{minipage}[c]{0.25\linewidth}
        \includegraphics[width=\linewidth]{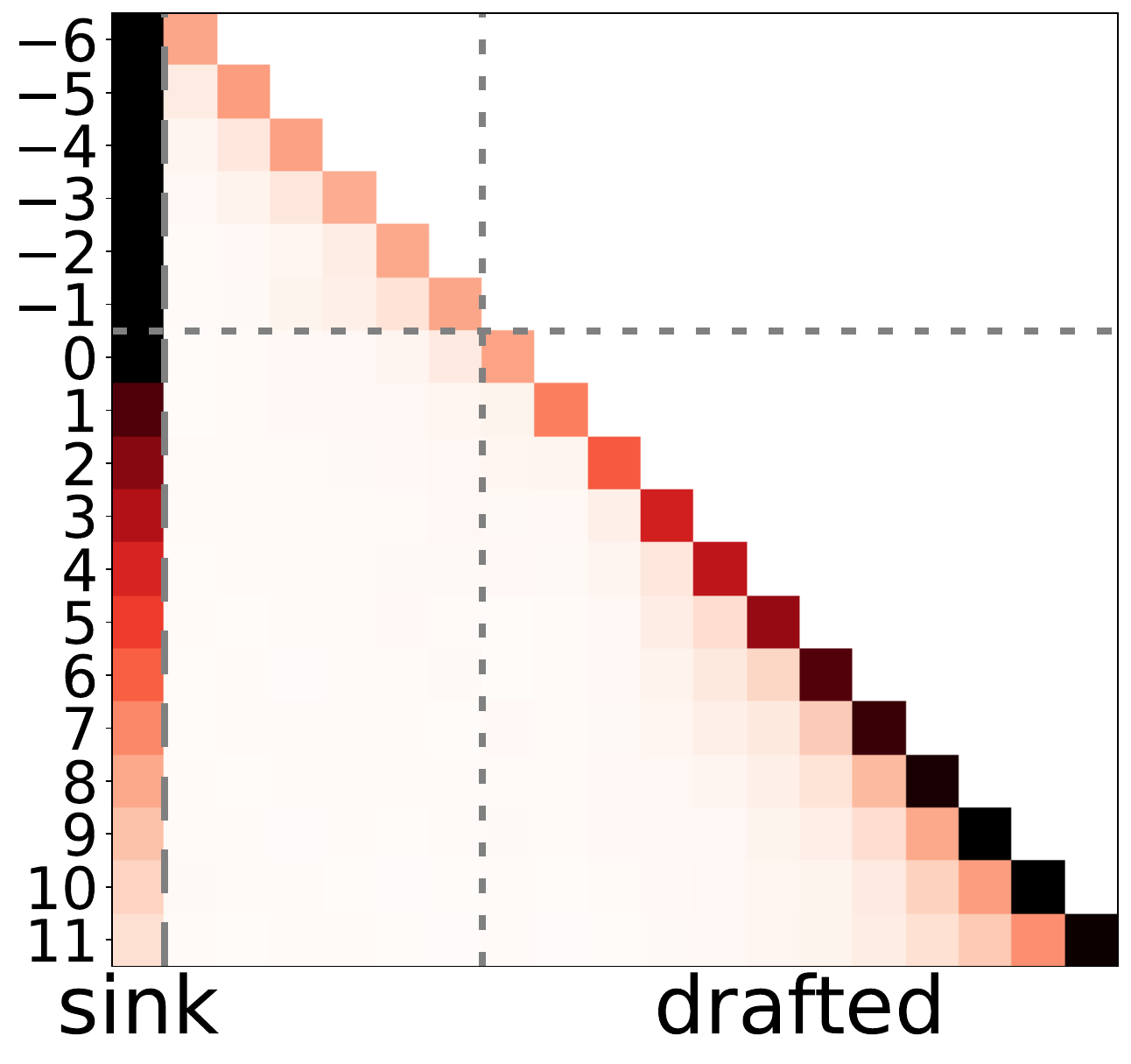}
    \end{minipage}%
    \hspace{0.02\linewidth}%
    \begin{minipage}[c]{0.55\linewidth}
        \resizebox{\linewidth}{!}{\begin{tikzpicture}[
    arr/.style={-{Stealth[length=3mm]}, thick},
]

% % === Triangle 1 ===
% \coordinate (t1-tl) at (0, 5);
% \coordinate (t1-bl) at (0, 0);
% \coordinate (t1-br) at (5, 0);

% \fill[blue!5] (t1-tl) -- (t1-bl) -- (t1-br) -- cycle;
% \draw[thick] (t1-tl) -- (t1-bl) -- (t1-br) -- cycle;

% \begin{scope}
%     \clip (t1-tl) -- (t1-bl) -- (t1-br) -- cycle;
%     \fill[red!80, opacity=0.7] (0, 0) rectangle (0.375, 5);
% \end{scope}

% === Triangle 2 ===
\coordinate (t2-tl) at (8.5, 5);
\coordinate (t2-bl) at (8.5, 0);
\coordinate (t2-br) at (13.5, 0);

\fill[blue!5] (t2-tl) -- (t2-bl) -- (t2-br) -- cycle;
\draw[thick] (t2-tl) -- (t2-bl) -- (t2-br) -- cycle;

\coordinate (t2-dash-l) at (7.9, 1.0);
\coordinate (t2-dash-r) at (13.3, 1.0);
\draw[dashed, thick, gray] (t2-dash-l) -- (t2-dash-r);

\begin{scope}
    \clip (t2-tl) -- (t2-bl) -- (t2-br) -- cycle;
    \fill[red!80, opacity=0.7] (8.5, 1.0) rectangle (8.875, 5);
\end{scope}

\begin{scope}
    \clip (t2-tl) -- (t2-bl) -- (t2-br) -- cycle;
    \shade[top color=red!80, bottom color=blue!5, opacity=0.6] (8.5, 0.25) rectangle (8.875, 1.0);
\end{scope}

\begin{scope}
    \clip (t2-bl) -- (8.5, 1.0) -- (12.5, 1.0) -- (t2-br) -- cycle;
    \shade[left color=blue!5, right color=red!100, opacity=0.7] (12.5, 0.01) rectangle (13.5, 1.0);
\end{scope}

% Label
% \node[font=\LARGE\bfseries, above=2.0cm of $(t1-tl)!0.5!(t1-br)$] {Verifier};
% \node[font=\LARGE\bfseries, above=2.0cm of $(t2-tl)!0.5!(t2-br)$] {Drafter};

% === Sink label for left triangle ===
% \draw[decorate, decoration={brace, amplitude=8pt, mirror}, thick]
    % ($(t1-tl) + (-0.5, 0)$) -- ($(t1-bl) + (-0.5, 0)$);
% \node[font=\LARGE\bfseries, left=0.8cm of $(t1-tl)!0.5!(t1-bl)$] {Sink};

\node[font=\large\bfseries, black, above=2pt of t2-dash-l, left=2pt, align=center] {Speculation\\starts};

% Overlay chevrons with increasing opacity
\draw[very thick, shorten >=2pt, -{Stealth[length=5mm, width=4mm]}] 
    ($(t2-bl) + (0.375, 0.875)$) to[bend right=15] ($(t2-br) + (-0.875, 0.25)$);

\path[postaction={decorate, decoration={markings,
    mark=at position 0.25 with {\arrow[black!50]{Stealth[length=5mm, width=4mm]}},
    mark=at position 0.5 with {\arrow[black!60]{Stealth[length=5mm, width=4mm]}},
    mark=at position 0.75 with {\arrow[black!70]{Stealth[length=5mm, width=4mm]}},
}}] 
    ($(t2-bl) + (0.375, 0.875)$) to[bend right=15] ($(t2-br) + (-0.875, 0.25)$);

\node[font=\LARGE\bfseries] at ($(t2-bl)!0.5!(t2-br) + (0, -1.0)$) {Attention Drift};

% Chart origin pushed right
\coordinate (chart-origin) at (15.0, 0.0);

\begin{scope}[shift={(chart-origin)}]
    % Axes
    \draw[thick, ->] (0, 0) -- (5, 0);
    \draw[thick, ->] (0, 0) -- (0, 4);

    % Vertical dashed line
    \draw[dashed, thick, gray] (3.5, 0) -- (3.5, 4);

    % Orange line (top, drops after dashed line)
    \draw[orange, line width=3pt] 
        (0, 3.2) -- (3.5, 2.9)
        to[out=-20, in=160] (4.8, 1.2);
    
    % Blue line (bottom, rises after dashed line)
    \draw[blue!50, line width=3pt] 
        (0, 1.5) -- (3.5, 1.75)
        to[out=20, in=200] (4.8, 3.5);
        
    % Title
    \node[font=\LARGE\bfseries] at (2.5, 4.6) {Amount of Attention};
    
    % Legend
    \draw[orange, line width=3pt] (0.5, -0.8) -- (1.2, -0.8);
    \node[font=\LARGE, right] at (1.2, -0.8) {Sink Token};
    
    \draw[blue!50, line width=3pt] (0.5, -1.5) -- (1.2, -1.5);
    \node[font=\LARGE, right] at (1.2, -1.5) {Recently Generated};
\end{scope}

\end{tikzpicture}}
    \end{minipage}
    \caption{
            \textbf{Attention drift.} During speculation, the drafter's attention moves from the prompt's sink token onto its own recently-generated tokens. \textit{Left:} Emergence of the attention sink demonstrated on drafter's attention heatmap, rows = query, columns = key; darker = higher attention. \textit{Center}: Graphical visualization of attention drift on a drafter. \textit{Right}: Attention per token position on $x$ axis, with speculated tokens to the right of the dashed line.
    }
    \label{fig:intro-attendrift}
\end{figure}
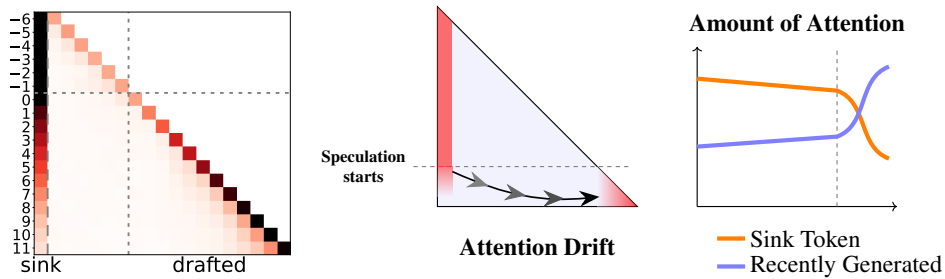

\section{Introduction}

% This story is from StreamingLLM (Attention Sink paper)

% > To understand the failure of window attention, we find an interesting phenomenon of autoregressive LLMs: a surprisingly large amount of attention score is allocated to the initial tokens, irrespective of their relevance to the language modeling task

Speculative decoding \cite{leviathan2023fast, chen2023accelerating} is a lossless acceleration technique for large language model (LLM) inference in which a lightweight \textit{drafter} predicts future tokens that are later verified by a larger \textit{target} or \textit{verifier} model. In practice, speculative decoding systems must operate under a wide range of deployment conditions, including different inference engines, context lengths, system prompts, and chat templates. Robustness under varying scenarios is critical to maintaining high acceptance rates and consistent inference speedups. However, recent works have shown that drafters often degrade substantially under challenging deployment scenarios, such as template perturbations and long-context inputs \cite{duoattention, longspec}. As a result, practitioners frequently retrain or specialize drafters for specific inference engines, prompts, or templates to maximize performance. Despite speculative decoding models being relatively cheap to train, this sensitivity points to an underlying issue. Our main contributions in this work are as follows:

\textbf{(i)} We identify a previously unreported phenomenon that helps explain this fragility, which we call \textbf{attention drift} (\Cref{fig:intro-attendrift}). As a drafter generates more tokens during speculation, its attention progressively shifts away from the sink toward recently generated tokens. We observe this phenomenon consistently across EAGLE-3 drafters~\cite{li2026eagle} and Multi-Token Prediction (MTP) heads~\cite{mtp}, suggesting that attention drift reflects a broader property of auto-regressive drafter designs.

% We first observe that drafters inherit the attention-sink behavior of their target models~\cite{streamingllm}.  Llama~\cite{grattafiori2024llama} drafters develop a sink on the first token, Qwen~\cite{yang2025qwen3} drafters on the second token, while GPT-oss~\cite{agarwal2025gpt} drafters exhibit no strong sink, mirroring the behavior of their corresponding targets. However, once drafting begins, the attention distribution changes substantially: attention gradually leaves the sink and increasingly concentrates on currently generated token. In models with strong sinks, sink attention weakens by approximately $3$--$5\times$ over the course of a speculation chain, and the majority of attention increasingly shifts toward the drafter's own \textit{speculated} tokens.

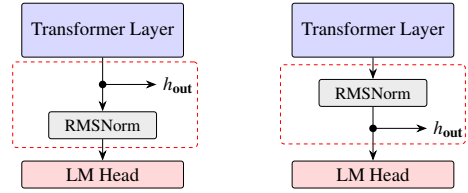
\begin{wrapfigure}{r}{0.4\linewidth}
  \centering
  \resizebox{\linewidth}{!}{\begin{tikzpicture}[
    scale=1.5, every node/.style={scale=1.5},
    block/.style={draw, rounded corners=2pt, minimum width=3cm, minimum height=0.5cm, font=\normalsize},
    smallblock/.style={draw, rounded corners=2pt, minimum width=2cm, minimum height=0.5cm, font=\small},
    tok/.style={font=\normalsize\bfseries, text height=1.5ex, text depth=0.25ex},
    arr/.style={-{Stealth[length=3mm]}, thick},
]

% Transformer layer
\node[block, fill=blue!15, minimum height=1.0cm] (tf) {Transformer Layer};

% Final RMSNorm
\node[smallblock, fill=gray!15, below=1.5cm of tf] (norm-out) {RMSNorm};
\draw[arr] (tf) -- (norm-out);

% LM Head
\node[block, fill=red!15, below=0.6cm of norm-out] (lmhead) {LM Head};
\draw[arr] (norm-out) -- (lmhead);

% h_out branches off the arrow between concat and transformer
\coordinate (h-tap) at ($(tf.south)!0.5!(norm-out.north)$);
\node[tok, right=1.5cm of h-tap] (h-out) {$h_{\text{out}}$};
\draw[arr] (h-tap) -- (h-out);
% Small dot to show the branch point
\fill (h-tap) circle (2pt);

% === Second block (h_out after RMSNorm) ===

\node[block, fill=blue!15, minimum height=1.0cm, right=3cm of tf] (tf2) {Transformer Layer};
\node[smallblock, fill=gray!15, below=0.6cm of tf2] (norm-out2) {RMSNorm};
\draw[arr] (tf2) -- (norm-out2);

\node[block, fill=red!15, below=1.5cm of norm-out2] (lmhead2) {LM Head};
\draw[arr] (norm-out2) -- (lmhead2);

% h_out after RMSNorm
\coordinate (h-tap2) at ($(norm-out2.south)!0.4!(lmhead2.north)$);
\node[tok, right=1.5cm of h-tap2] (h-out2) {$h_{\text{out}}$};
\draw[arr] (h-tap2) -- (h-out2);
\fill (h-tap2) circle (2pt);

% Labels
% \node[font=\large\bfseries, below=1.0cm of $(lmhead)$] {Original};
% \node[font=\large\bfseries, below=1.0cm of $(lmhead2)$] {Ours};

% Red dashed circles
\node[draw=red, dashed, thick, rounded corners=3pt, inner sep=1pt, 
      fit=(h-tap)(h-out)(norm-out), scale=0.8,
      xshift=-0.44cm] (circle-left) {};
\node[draw=red, dashed, thick, rounded corners=3pt, inner sep=1pt,
      fit=(h-tap2)(h-out2)(norm-out2), scale=0.8,
      yshift=0.15cm, xshift=-0.5cm] (circle-right) {};

% Double-sided arrow between them
% \coordinate (arrow-y) at ($(circle-left.east)!0.5!(circle-right.west)$);
% \draw[red, line width=8pt, {Stealth[length=6mm, width=8mm]}-{Stealth[length=6mm, width=8mm]}, shorten <=10pt, shorten >=10pt] 
%     (circle-left.east |- arrow-y) -- (circle-right.west |- arrow-y);
    
\end{tikzpicture}}
  \caption{Overview of Pre-norm (Left) and proposed Post-norm (Right) architecture.}
  \label{fig:into-prepost}
\end{wrapfigure}

\textbf{(ii)} To understand this behavior, we analyze the hidden-state dynamics of speculative drafters and find that the unnormalized residual connection between speculation steps causes hidden-state magnitudes to grow monotonically with chain depth, resembling additional transformer layers stacked on top of the verifier. The drafter implicitly learns a depth-dependent refinement process instead of a stable autoregressive token predictor, making it sensitive to template changes and long contexts (\Cref{sec:normpos}).

\textbf{(iii)} Motivated by this observation, we introduce a simple architectural modification based on \textit{post-normalization} (\Cref{fig:into-prepost}) combined with a normalization before hidden-state fusion, that prevents hidden-state magnitude growth and stabilizes the drafting process. These changes substantially reduce attention drift and improve acceptance length over the current state-of-the-art pre-norm EAGLE3 architecture by up to $2\times$ under template perturbation, $1.18\times$ on long-context tasks, and $1.10\times$ across seven standard benchmarks spanning multi-turn chat, reasoning, coding, and mathematics. We further show that post-norm drafters generalize to inference depths beyond those seen during training, enabling shorter train-time-test depths and reducing training cost.

% 1. We identify \textbf{Attention Drift}, a previously unreported phenomenon in speculative decoding drafters where attention progressively shifts toward recently generated tokens during drafting.

% 2. We trace attention drift to the unnormalized residual path between chain steps, showing that commonly used pre-norm drafters implicitly behave like additional transformer layers stacked on top of the target rather than as standalone autoregressive predictors.

% 3. We propose two simple architectural interventions: 1) post-norm on the chain residual, 2) per-stream RMS normalization before the fully connected (FC) layer that fuses hidden states from the target and drafter. We demonstrate that these changes substantially improve robustness and acceptance length over the current state-of-the-art (SOTA): pre-norm EAGLE-3 \cite{li2026eagle}, by up to $100\%$ under template perturbation, $18\%$ on long-context tasks, and $10\%$ on seven standard benchmarks spanning multi-turn chat, math, and coding.

% \hy{Perhaps give a figure here to show post-norm vs. pre-norm.}

% \hy{I would suggest expanding the phenomenon description in more details in the introduction.}

\section{Preliminaries}

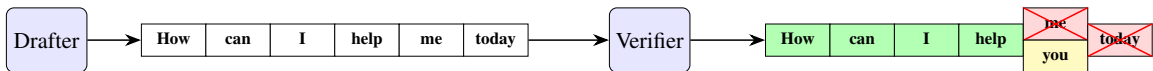
\begin{figure}[b!]
    \resizebox{1.0\linewidth}{!}{\begin{tikzpicture}[
    tok/.style={
        font=\small\bfseries,
        minimum width=1.2cm,
        minimum height=0.6cm,
        draw,
        rounded corners=0pt,
        align=center,
        inner sep=3pt,
    },
    box/.style={draw, rounded corners, minimum width=1.5cm, minimum height=1.2cm, font=\small, fill=blue!10},
    arr/.style={-{Stealth[length=3mm]}, thick},
]

% Input tokens (horizontal)
\foreach \word [count=\i] in {How, can, I, help, me, today} {
    \node[tok, fill=white] (in-\i) at (\i*1.2, 0) {\strut\word};
}

% Drafter model
\node[box, left=1.0cm of in-1] (drafter) {\large Drafter};

% Target model
\node[box, right=1.5cm of in-6] (target) {\large Verifier};

% Output tokens (horizontal)
\foreach \word/\col [count=\i] in {How/green!30, can/green!30, I/green!30, help/green!30} {
    \node[tok, fill=\col] (out-\i) at ($(target.east) + (0.8 + \i*1.2, 0)$) {\strut\word};
}

% "me" crossed out on top
\node[tok, fill=red!15] (out-5-old) at ($(target.east) + (0.8 + 5*1.2, 0.3)$) {\strut me};
\draw[red, thick] (out-5-old.south west) -- (out-5-old.north east);
\draw[red, thick] (out-5-old.north west) -- (out-5-old.south east);

% "you" replacement below
\node[tok, fill=yellow!30] (out-5-new) at ($(target.east) + (0.8 + 5*1.2, -0.3)$) {\strut you};

% "today" crossed out
\node[tok, fill=red!15] (out-6) at ($(target.east) + (0.8 + 6*1.2, 0)$) {\strut today};
\draw[red, thick] (out-6.south west) -- (out-6.north east);
\draw[red, thick] (out-6.north west) -- (out-6.south east);

% Arrows
\draw[arr] (drafter.east) -- (in-1.west);
\draw[arr] (in-6.east) -- (target.west);
\draw[arr] (target.east) -- (out-1.west);

\end{tikzpicture}}
    \caption{Verification phase: green tokens are accepted, yellow is resampled and red ones are rejected.}
    \label{fig:verify}
\end{figure}

\textbf{Speculative Decoding} alternates between two stages (\Cref{fig:verify}). In the drafting phase, the drafter auto-regressively generates a sequence of $k$ candidate future tokens, where $k$ is the \textit{speculation depth}. In the verification phase, the target model evaluates the drafted sequence in a single forward pass and applies rejection sampling against its own token distribution; the longest valid prefix is accepted. A key efficiency metric is the \textit{acceptance length ($\tau$)}, defined as the average number of drafted tokens accepted per verification round. Higher acceptance lengths generally translate directly into larger end-to-end inference speedups for a fixed drafter overhead.

\textbf{EAGLE} \citep{eagle1} and its successors EAGLE-2/EAGLE-3 \citep{li2024eagle, li2026eagle} feed the target's fused internal hidden states through a fully-connected projection (\textsc{fc} in \Cref{fig:postnorm-arch}) into a single pre-norm transformer decoder layer~\cite{vaswani2017attention} that serves as the drafter (\Cref{fig:into-prepost} Left). The drafter is trained with a token-level cross-entropy loss against the frozen target's predicted distribution, with a fixed number of speculation steps (the train-time-test depth, TTT). EAGLE-3, our focus, is the dominant auto-regressive drafter design in production engines such as vLLM \cite{vllm} and SGLang \cite{sglang}.

\textbf{Attention Sinks.} An attention sink \citep{streamingllm} is a token (typically near the start of a sequence) that absorbs disproportionately large attention during inference. Sinks are widely observed in modern LLMs and are believed to act as stable anchors that stabilize attention, particularly under long-context.

\section{Attention Drift}

\begin{figure}[h]
    \centering
    \begin{subfigure}{0.48\linewidth}
        \centering
        \includegraphics[width=\linewidth]{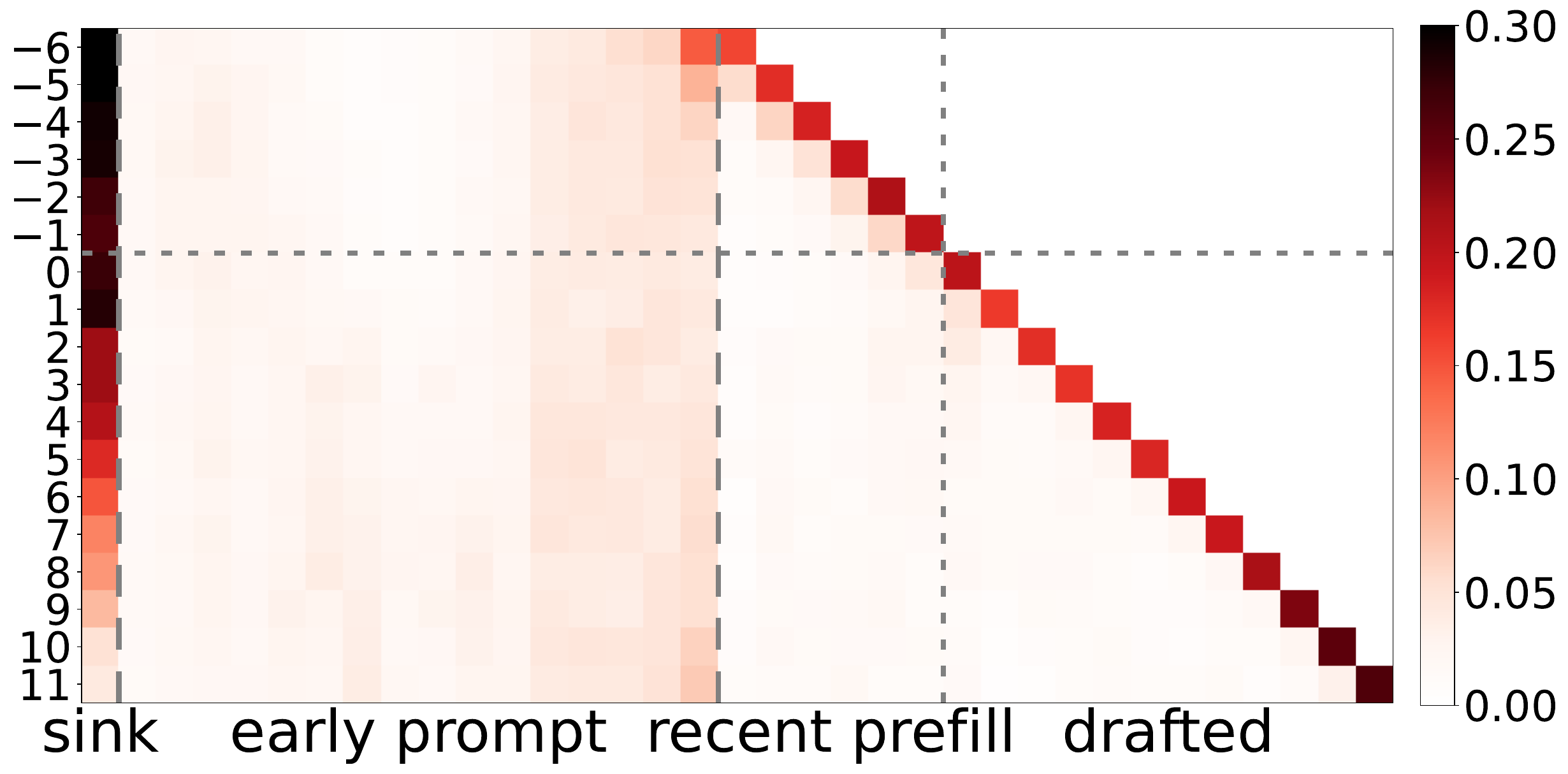}
        \caption{Qwen3.5 35B}
    \end{subfigure}
    \hfill
    \begin{subfigure}{0.48\linewidth}
        \centering
        \includegraphics[width=\linewidth]{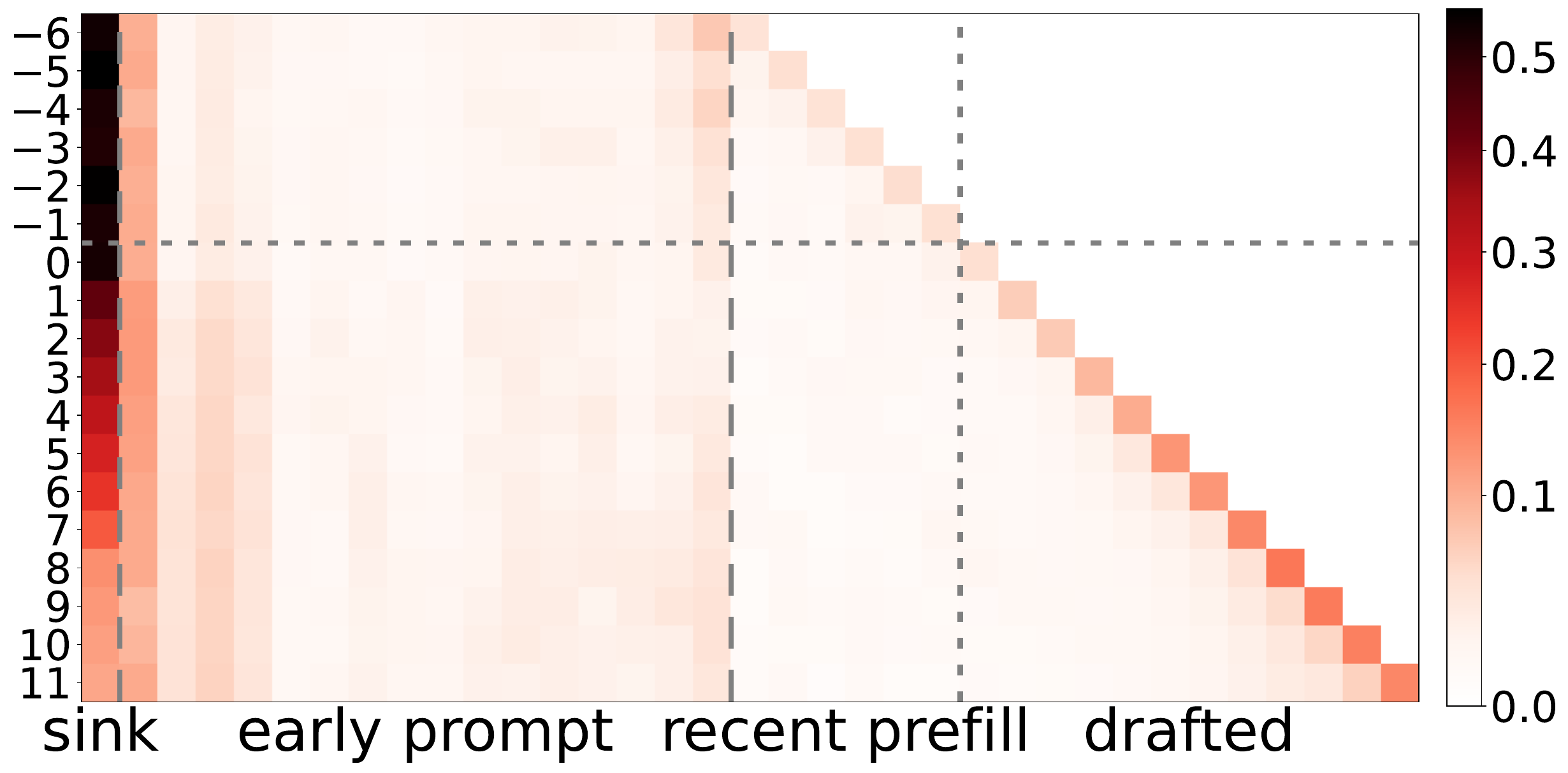}
        \caption{Llama3 8B.}
    \end{subfigure}
    \caption{Attention heatmaps for visualizing attention drift on EAGLE-3 drafters. Aggregated over 200+ samples from varied prompts and sequence positions.}
    \label{fig:drifts}
\end{figure}

Looking at speculative decoding models' attention, a pattern emerges: \textbf{whenever the target model develops a sink, the drafter develops a sink at exactly the same place}. This observation is consistent across \textbf{MTP heads} and \textbf{EAGLE-3} drafters. Upon close inspection of three model families with different attention patterns, we observed Llama's sink is on the first token, Qwen's on the second, and GPT-oss has no sink. In every case, the drafter matched its target.

% We begin by characterizing the attention behavior of autoregressive speculative drafters during generation. Although our primary analysis focuses on \textbf{EAGLE3} drafters due to their state-of-the-art performance and widespread deployment in mod  ern inference systems~\cite{li2026eagle}, we observe qualitatively similar behavior across other speculative decoding architectures, including \textbf{MTP heads} (\cref{sec:mtp-heads}). Across all model families we studied, \textbf{whenever the target model develops a sink, the drafter develops a sink at exactly the same place}: Llama drafters concentrate strongly on the first token, Qwen drafters on the second token, while GPT-oss models exhibit little or no sink behavior. Visualizations and further discussions in \cref{some appendix}.

The drafter attention resembles the verifier attention as the drafter consumes verifier hidden states. However, once drafting begins, the attention distribution progressively changes as the drafter transitions from consuming verifier hidden states to consuming its own generated hidden states. During this process, attention increasingly shifts away from the sink and toward recently generated tokens.

\begin{figure}[h]
    \centering
    \includegraphics[width=0.8\linewidth]{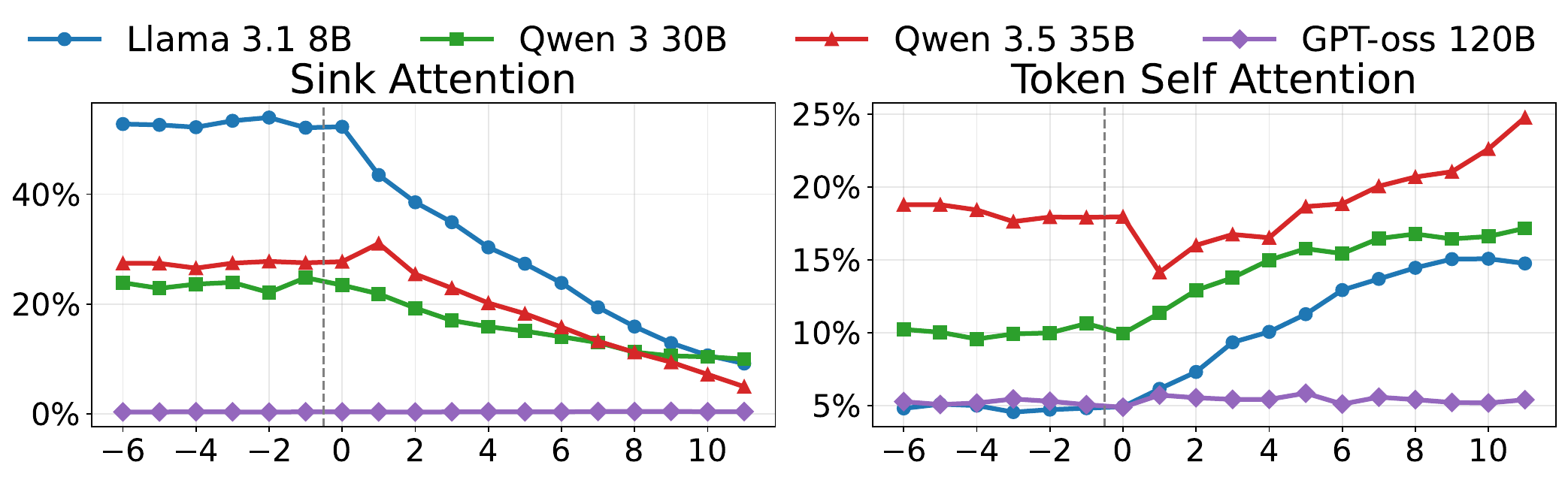}
        \caption{Percentage of attention concentrated on Sink token (Left) and Latest Token (Right) on EAGLE-3 drafters.}
    \label{fig:drifts-line}
\end{figure}

We refer to this phenomenon as \textbf{attention drift}, visualized in \Cref{fig:drifts,fig:drifts-line} across multiple EAGLE-3 drafters. In models with strong sinks, the sink progressively weakens during drafting while attention mass concentrates on recently drafted tokens. The drafter therefore operates in two modes: attending to the verifier's hidden states at pre-fill and to its own hidden states during generation, with a gradual learned transition between them. We hypothesize that this transition makes the drafter \textbf{more fragile to out-of-distribution inputs}, most notably in long context and at deeper speculation steps.

\vspace{10pt}

% To quantify this behavior, we measure the total attention mass assigned to 1) the sink token and 2) the most recently generated token across drafting steps, summarized in \Cref{fig:drifts-line} for four representative drafter families. We observe two consistent trends.

% 1. \textbf{Attention leaves the sink} and drops around $2$--$3\times$ during speculation. Llama, which concentrates heavily on BoS at pre-fill ($0.47$), loses almost all of it by $k{=}8$ ($0.08$); Qwen's sink on token $1$ halves ($0.24 \to 0.12$) (\cref{fig:drifts,fig:drifts-line}); GPT-oss-120B (\cref{fig:attngpt}) carries no appreciable sink at any step.

% 2. \textbf{Total attention on most recently drafted tokens grows with drafting depth} and the drafter spends an increasing fraction of its attention budget looking at its own predictions.

\newpage

Here, we examine MTP Heads~\cite{mtp} as a cross-architecture check on whether attention drift is specific to EAGLE-3 or a more general property of drafters that consume their own previous outputs. MTP Heads are auxiliary prediction heads jointly trained with the target model during pretraining, sharing the target's LM head and predicting multiple positions ahead. This differs from EAGLE-3, which is trained post-hoc as a separate drafter on top of a frozen target with its own LM head.

\begin{figure}[h!]
        \centering
        \includegraphics[width=0.48\linewidth]{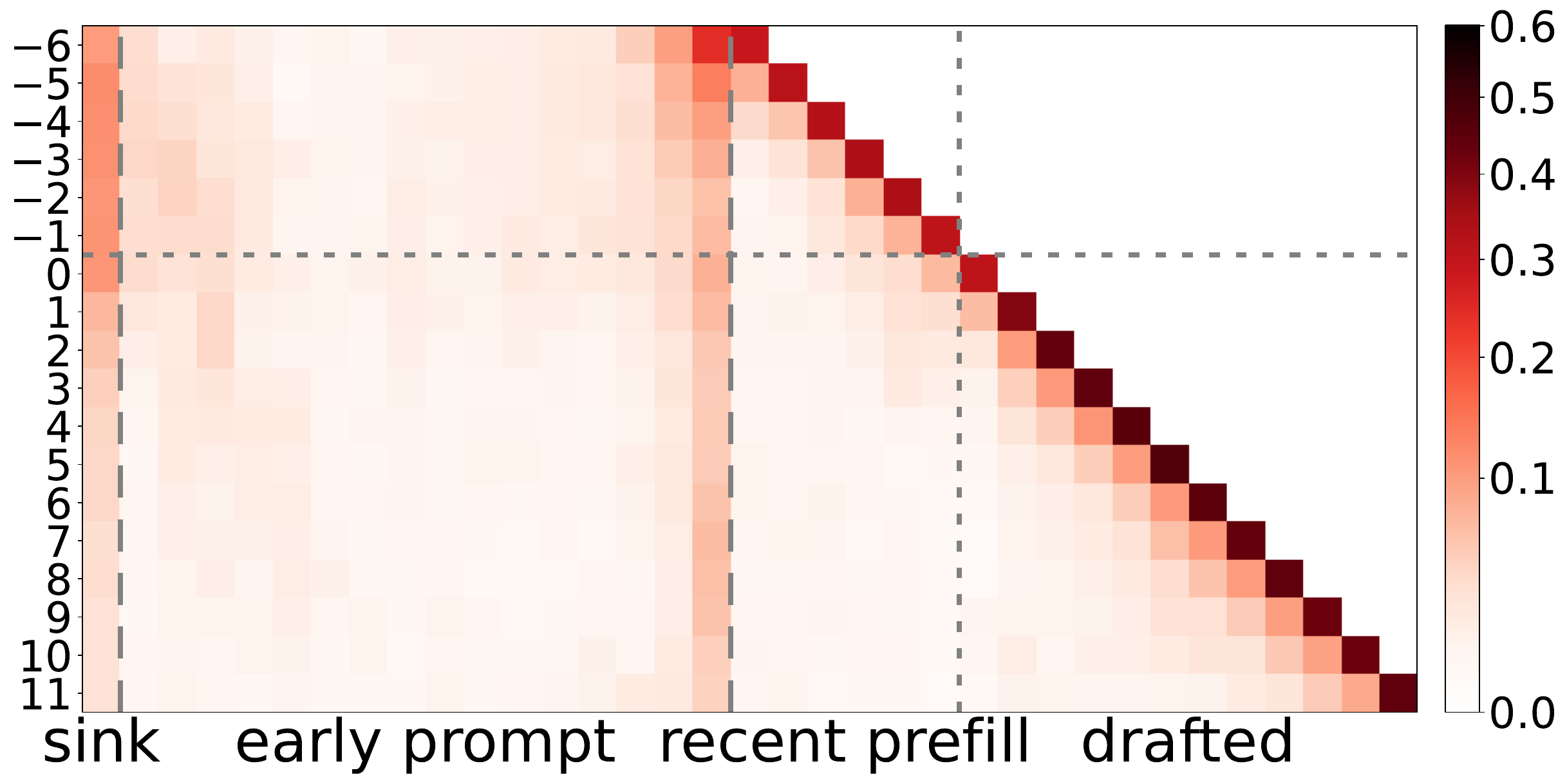}
        \label{fig:attnmtpqwen}
        \caption{Attention Drift on Qwen3.5 9B MTP head}
\end{figure}

\begin{figure}[h!]
    \centering
    \includegraphics[width=0.8\linewidth]{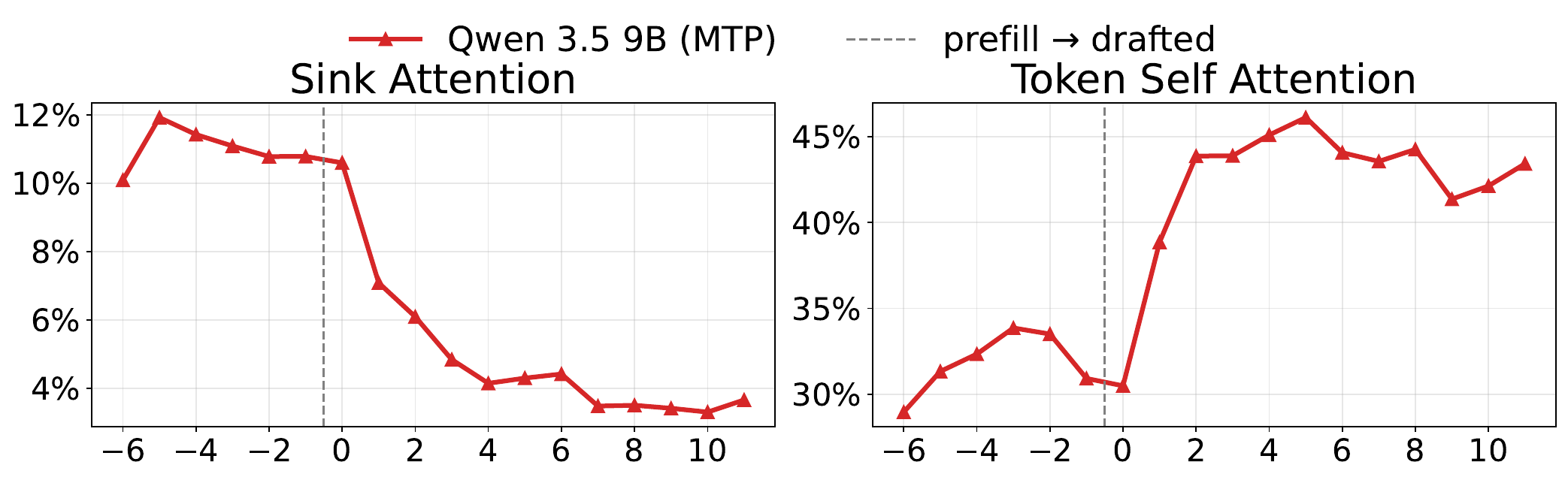}
    \caption{Sink and drafted token self attention on Qwen3.5 9B MTP heads (MT-Bench, 80 prompts).}
    \label{fig:mtpattn}
\end{figure}

We inspect the MTP head of Qwen3.5 9B (\Cref{fig:attnmtpqwen}), a single transformer layer that follows the target model’s attention architecture and reuses the same weights across consecutive speculation steps. During MTP speculation, we observe similar trends (\Cref{fig:mtpattn}): attention to the sink token decreases substantially, while attention to the most recently drafted token increases.

\begin{wrapfigure}{r}{0.4\linewidth}
  \centering
  \includegraphics[width=\linewidth]{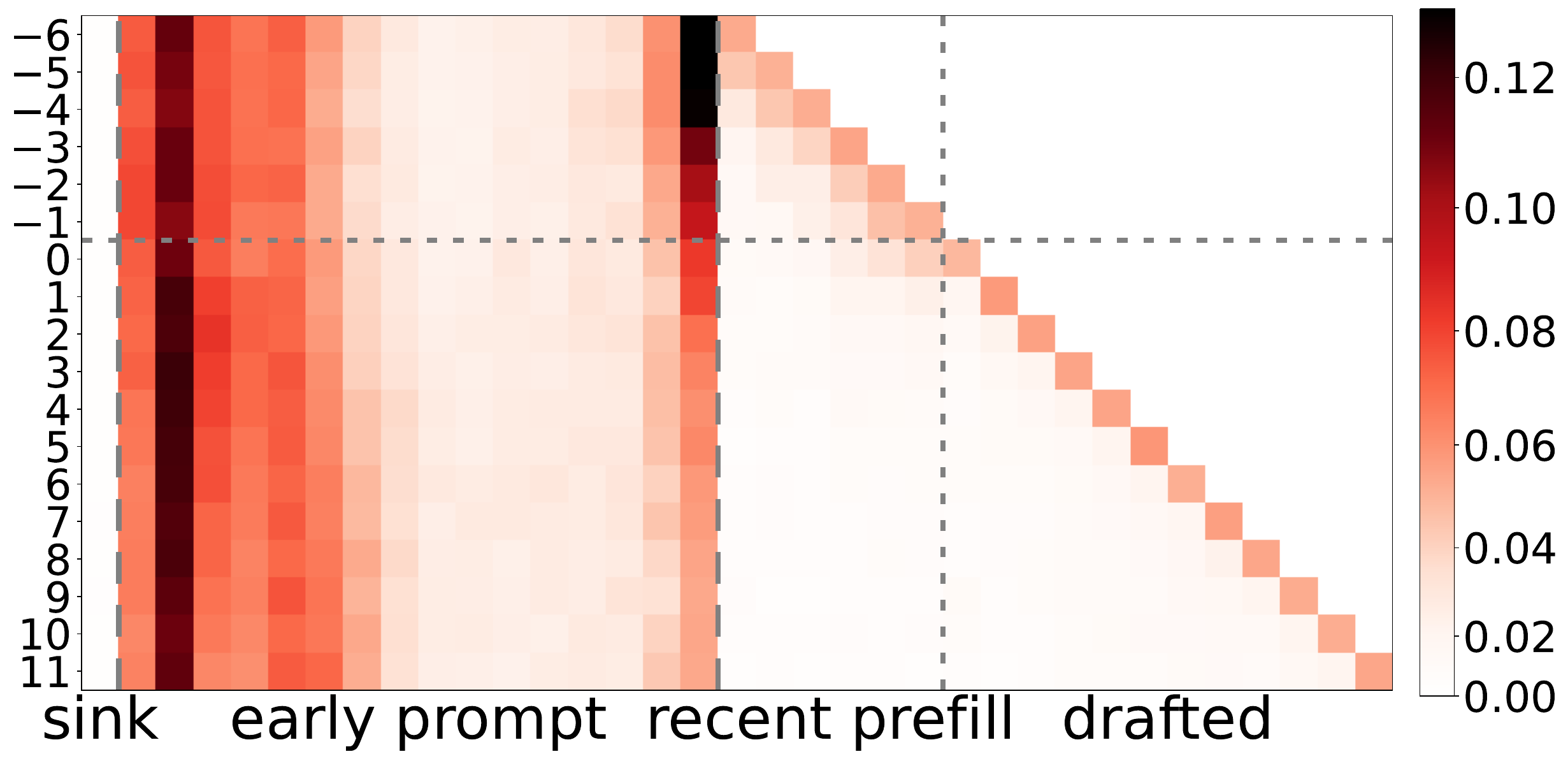}
  \caption{Attention Drift on GPT-oss 120B.}
  \label{fig:attngpt-oss}
\end{wrapfigure}

\textbf{What if verifier doesn't have an attention sink?} Some recent models, such as Qwen3-Next and GPT-oss, are designed to suppress attention sinks. Qwen3-Next uses gated attention, applying a per-head sigmoid gate to the SDPA output so that heads can multiplicatively suppress their contribution. GPT-oss instead introduces a per-head learnable bias logit in the softmax denominator, giving each head an explicit "attend to nothing" option that absorbs excess attention mass. Inspecting a GPT-oss drafter (\Cref{fig:attngpt-oss}), we observe no visible sink on the first token, but a weak yet consistent attention concentration on recurring template tokens. This suggests that sink-like behavior can arise not only from architectural inductive bias but also from repeated special tokens or template markers, especially in the absence of a dedicated architectural sink mechanism in the verifier.

% Interestingly, GPT-oss models exhibit comparatively weak drift despite lacking a strong attention sink. This suggests that attention sinks amplify the phenomenon but are not its sole cause.

% These observations suggest that speculative drafters progressively transition away from prompt-conditioned behavior as drafting depth increases. In the next section, we investigate the hidden-state dynamics underlying this shift and show that the phenomenon is closely tied to magnitude accumulation in the recursive residual pathway of standard pre-norm drafters.

\section{What Causes Attention Drift?}

To understand why attention drift occurs and what it reveals about the underlying model, we trained drafters from three different verifier model families. We primarily evaluate on MT-Bench~\cite{mtbench}, eight categories of multi-turn instructions allows us to evaluate across diverse prompts. All architectural changes explored in this section required re-training of the drafter (details in Appendix \ref{ap:training}).

\subsection{Hidden State Magnitudes}

Attention drift occurs as we move away from the target's hidden states towards consuming the drafter's own hidden states during generation. Therefore we first inspect the hidden states of the verifier and drafter. In \Cref{tab:hidden-magnitudes}, the magnitudes are compared between the verifier's hidden states, verifier's fused hidden state and the drafter's own hidden states at different speculation steps, using root mean square (RMS), $\|x\|_2 / \sqrt{d}$. We notice several patterns described next.

\begin{table}[h]
\centering
\caption{Hidden state magnitudes (RMS) across model families, averaged over 80 MT-Bench prompts and $k{=}8$ speculation rounds. Hidden states from the verifier's low, middle, and high layers ($h_{low}$, $h_{mid}$, $h_{high}$); the fused representation of the verifier's hidden states ${h_{FC}}$; and the drafter's output hidden states at speculation steps 1, 2, 3, and 8 ($h^*_1$ through $h^*_8$)}
\begin{tabular}{lccc|c|cccc}
\toprule
& \multicolumn{3}{c|}{Verifier hidden states} & & \multicolumn{4}{c}{Drafter output $h^*_t$} \\
\cmidrule(lr){2-4} \cmidrule(lr){6-9}
Target Model & $h_{\text{low}}$ & $h_{\text{mid}}$ & $h_{\text{high}}$ & ${h_{FC}}$ & $k{=}1$ & $k{=}2$ & $k{=}3$ & $k{=}8$ \\
\midrule
Llama 3.1 8B   & 0.56 & 0.58 & 0.78 & 12.46 & 3.92 & 4.87 & 5.86 & 14.02 \\
Qwen 3 30B     & 0.33 & 2.17 & 2.71 & 2.56  & 1.00 & 1.13 & 1.25 & 1.67  \\
Qwen 3.5 35B   & 0.03 & 0.12 & 0.21 & 0.89  & 3.47 & 4.03 & 4.42 & 5.92  \\
GPT-oss 20B    & 3    & 52   & 324  & 96    & 87   & 89   & 95   & 138   \\
GPT-oss 120B   & 3    & 163  & \textbf{1455} & 583 & 497 & 511 & 537 & \textbf{647} \\
\bottomrule
\end{tabular}
\label{tab:hidden-magnitudes}
\end{table}

\textbf{Observation 1: Hidden-state magnitudes mismatch.} The hidden states of the verifier, target, and the fully connected layer have substantially different magnitudes. This is attributed to the positioning of the RMSNorm in the drafter and a norm component being shared between (${h_{FC}}$) and the drafter ($h^*$) shown in \Cref{fig:postnorm-arch}. Despite the hidden state magnitudes being vastly different between the target and drafter, we did not see a correlation between the magnitude of hidden states and acceptance rates for models, hinting us that the model learns to compensate for it.

\textbf{Observation 2: The verifier fused representation, ${h_{FC}}$, is imbalanced.} All three target families we explored use the pre-norm architecture: RMSNorm is applied inside each transformer block (before attention and MLP), but never to the residual stream itself. As such, each layer's magnitudes accumulate monotonically through depth (prenorm dilution \cite{attentionresiduals}), and $\|h_1\| < \|h_2\| < \|h_3\|$ holds for every row of Table~\ref{tab:hidden-magnitudes}. The target's final pre-LM-head RMSNorm normally absorbs this growth when producing logits, but EAGLE-3 uses verifier's internal states captured before this norm ($h_{low}$, $h_{mid}$, $h_{high}$). Because of this $h_{high}$ dominates the $h_{FC}$ signal with its larger magnitude. Thus in future sections \textbf{we add RMSNorm layers before each target hidden stream} used as input to generate ${h_{FC}}$ to prevent this imbalance and generate a more stable verifier representation.

\textbf{Observation 3: Magnitude growth across speculation depth.} The drafter-generated hidden states grow monotonically with speculation depth. Across all model families, $h^*_k$ accumulates magnitude away from the distribution the drafter was trained on at step $0$. This means that the drafter does not operate on a depth-invariant hidden-state distribution. Instead, each speculation step changes the scale of the representation consumed by the next step.

\begin{insight}
\textbf{How is drift and magnitude related?} This scaling magnitude across speculation depth is important in computing attention. Attention depends on computing query-key similarities. As the drafter generates deeper into the speculation chain, its current hidden state increasingly resembles recently generated drafter states rather than verifier-conditioned prompt states. The empirical consequence, quantified in the next section, is a \emph{redistribution} of attention mass away from the original input and onto the drafter's own predictions.
\end{insight}

\subsection{Norm Position Changes: Layer-stacking vs Autoregression}
\label{sec:normpos}

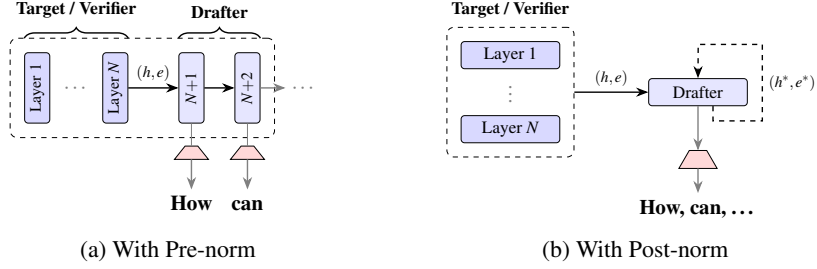
\begin{figure}[h]
  \centering
  \begin{subfigure}[t]{0.4\linewidth}
    \centering
    \vbox to 3cm{\vfill\resizebox{!}{3cm}{\begin{tikzpicture}[
    vlayer/.style={draw, rounded corners=2pt, minimum width=0.5cm, minimum height=1.4cm, fill=blue!15},
    vdrafter/.style={draw, rounded corners=2pt, minimum width=0.5cm, minimum height=1.4cm, fill=blue!10},
    vlabel/.style={font=\small, rotate=90, anchor=center},
    tok/.style={font=\large\bfseries},
    arr/.style={-{Stealth[length=2.5mm]}, thick},
    dots/.style={font=\large, text=gray},
]
  % --- Verifier layers ---
  \node[vlayer] (L-l1) {};
  \node[vlabel] at (L-l1) {Layer 1};
  \node[dots, right=0.15cm of L-l1] (L-dd) {$\cdots$};
  \node[vlayer, right=0.15cm of L-dd] (L-lN) {};
  \node[vlabel] at (L-lN) {Layer $N$};

  % --- Drafter layers ---
  \node[vdrafter, right=1.0cm of L-lN] (L-dr1) {};
  \node[vlabel] at (L-dr1) {$N{+}1$};
  \node[vdrafter, right=0.6cm of L-dr1] (L-dr2) {};
  \node[vlabel] at (L-dr2) {$N{+}2$};
  \node[dots, right=0.5cm of L-dr2] (L-ddots) {$\cdots$};

  \draw[arr] (L-lN.east) -- node[above, font=\small] {$(h,e)$} (L-dr1.west);
  \draw[arr] (L-dr1.east)  -- (L-dr2.west);
  \draw[arr, gray] (L-dr2.east) -- (L-ddots.west);

  % --- Drafter brace (anchored to slabs, horizontal) ---
  \coordinate (br-l) at ([yshift=8pt]L-dr1.north west);
  \coordinate (br-r) at ([yshift=8pt]L-dr2.north east);
  \draw[decorate, decoration={brace, amplitude=6pt}, thick] (br-l) -- (br-r);
  \node[font=\bfseries, above=8pt of $(br-l)!0.5!(br-r)$] {Drafter};

  % --- LM heads + tokens below, all on one baseline ---
  \foreach \i/\word in {1/How, 2/can} {
    \node[minimum width=0.45cm, minimum height=0.25cm, below=0.45cm of L-dr\i] (L-lm\i) {};
    \fill[red!15]
      ([xshift=3pt]L-lm\i.north west) -- ([xshift=-3pt]L-lm\i.south west) --
      ([xshift=3pt]L-lm\i.south east) -- ([xshift=-3pt]L-lm\i.north east) -- cycle;
    \draw[thin]
      ([xshift=3pt]L-lm\i.north west) -- ([xshift=-3pt]L-lm\i.south west) --
      ([xshift=3pt]L-lm\i.south east) -- ([xshift=-3pt]L-lm\i.north east) -- cycle;
    \draw[thin, gray] (L-dr\i.south) -- (L-lm\i.north);
    \draw[arr, gray] (L-lm\i.south) -- ++(0,-0.6cm);
  }
  \node[tok] at (L-lm1 |- 0,0) {}; % placeholder for alignment reference
  \coordinate (tok-y) at ([yshift=-1.0cm]L-lm1.south);
  \node[tok, anchor=base] at (L-lm1 |- tok-y) (L-t1) {\strut How};
  \node[tok, anchor=base] at (L-lm2 |- tok-y) (L-t2) {\strut can};

  \begin{pgfonlayer}{background}
    \node[draw, dashed, rounded corners=4pt, inner sep=8pt,
          fit=(L-l1)(L-lN)(L-dr1)(L-dr2), fill=white] (L-vbox) {};
  \end{pgfonlayer}

% --- Target / Verifier (anchored to slabs, horizontal) ---
  \coordinate (br-l) at ([yshift=8pt]L-l1.north west);
  \coordinate (br-r) at ([yshift=8pt]L-lN.north east);
  \draw[decorate, decoration={brace, amplitude=6pt}, thick] (br-l) -- (br-r);
  \node[font=\bfseries, above=8pt of $(br-l)!0.5!(br-r)$] {Target / Verifier};

  % \node[font=\bfseries, above=2pt of L-vbox.north] {Target / Verifier};

\end{tikzpicture}}\vfill}
    \caption{With Pre-norm}
    \label{fig:stacking}
  \end{subfigure}
  \begin{subfigure}[t]{0.4\linewidth}
    \centering
    \vbox to 3cm{\vfill\resizebox{!}{3cm}{\begin{tikzpicture}[
    block/.style={draw, rounded corners=2pt, minimum width=2cm, minimum height=0.55cm},
    drafterblock/.style={draw, rounded corners=2pt, minimum width=2cm, minimum height=0.55cm, fill=blue!10},
    tok/.style={font=\large\bfseries},
    arr/.style={-{Stealth[length=2.5mm]}, thick},
    feedback/.style={-{Stealth[length=2.5mm]}, thick, dashed},
    dots/.style={font=\large, text=gray},
]
  \node[block, fill=blue!15] (R-l1) {Layer 1};
  \node[dots, below=0.0cm of R-l1] (R-dd) {\vdots};
  \node[block, fill=blue!15, below=0.15cm of R-dd] (R-lN) {Layer $N$};
  \begin{pgfonlayer}{background}
    \node[draw, dashed, rounded corners=4pt, inner sep=8pt,
          fit=(R-l1)(R-lN), fill=white] (R-vbox) {};
  \end{pgfonlayer}
  \node[font=\bfseries, above=2pt of R-vbox.north] {Target / Verifier};

  \node[drafterblock, right=1.5cm of R-vbox] (R-dr) {Drafter};

  \draw[arr] (R-vbox.east) -- node[above, font=\small] {$(h,e)$} (R-dr.west);

  % Loop: bottom -> right -> up -> left -> bottom
  \draw[feedback]
    ([xshift=-20pt]R-dr.south east)
    -- ++(0,-0.30)
    -- ++(1.0,0)
    -- node[right, font=\small] {$(h^*,e^*)$} ++(0,1.55)
    -- ++(-1.3,0)
    -- ([xshift=0pt]R-dr.north);

  \node[minimum width=0.55cm, minimum height=0.35cm, below=0.9cm of R-dr] (R-lm) {};
  \fill[red!15]
    ([xshift=3pt]R-lm.north west) -- ([xshift=-3pt]R-lm.south west) --
    ([xshift=3pt]R-lm.south east) -- ([xshift=-3pt]R-lm.north east) -- cycle;
  \draw
    ([xshift=3pt]R-lm.north west) -- ([xshift=-3pt]R-lm.south west) --
    ([xshift=3pt]R-lm.south east) -- ([xshift=-3pt]R-lm.north east) -- cycle;
  \draw[arr, gray] (R-dr.south) -- (R-lm.north);

  \node[tok, below=0.5cm of R-lm] (R-t) {How, can, \dots};
  \draw[arr, gray] (R-lm.south) -- (R-t);
\end{tikzpicture}}\vfill}
    \caption{With Post-norm}
    \label{fig:autoregressive}
  \end{subfigure}
  \caption{Two views of the EAGLE3 drafter at chain depth $>1$. \textbf{With the standard pre-norm structure, the model acts more like additional layers stacked onto the target model, whereas post-norm acts more like an independent auto-regressive drafter that accepts its own output.}}
  \label{fig:drafter-modes}
\end{figure}

The monotonic growth of the drafter hidden states suggests a simple interpretation of what a drafter \emph{learns to be}. In a standard pre-norm drafter, the drafter's chain hidden states $h^*_1, h^*_2, \dots, h^*_k$ accumulate through an unnormalized residual path, causing the hidden states to monotonically grow with speculation depth. This makes the drafter behave less like an independent autoregressive model and more like a sequence of additional transformer layers stacked on top of the target. At speculation step $k$, the drafter effectively approximates what would happen if the target had $N, N{+}1, \dots, N{+}k$ layers, rather than acting as a standalone depth-invariant model that autoregressively predicts the next token. The model takes the target's representations and \emph{keeps refining it} with another layer of attention+MLP compute, one per speculation step (\Cref{fig:drafter-modes}).

%\textbf{So what happens if we feed drafter with the hidden state that is Post-norm into itself?} 

\begin{wrapfigure}{r}{0.52\linewidth}
    \centering
    \begin{subfigure}[h]{1.0\linewidth}
        \centering
        \small
        \small  % or \scriptsize for even smaller
        \begin{tabular}{l|cc|cc|c}
            \toprule
            Drafter / TTT & $h^*_{1}$ & $h^*_{8}$ & $k{=}1$& $k{=}16$ & Accept. \\
            \midrule
            Pre-norm / 8  & 4.15 & 13.10 & 0.91 & 0.50 & \textbf{7.12} \\
            Pre-norm / 2  & 9.76 & 44.52 & \textbf{0.93} & 0.00 & 2.64 \\
            Post-norm / 2 & 1.22 & 1.30  & \textbf{0.93} & \textbf{0.53} & 5.05 \\
            \bottomrule
        \end{tabular}
        \vspace{20pt}
    \end{subfigure}
    \begin{subfigure}[h]{1.0\linewidth}
        \centering
        \includegraphics[width=\linewidth]{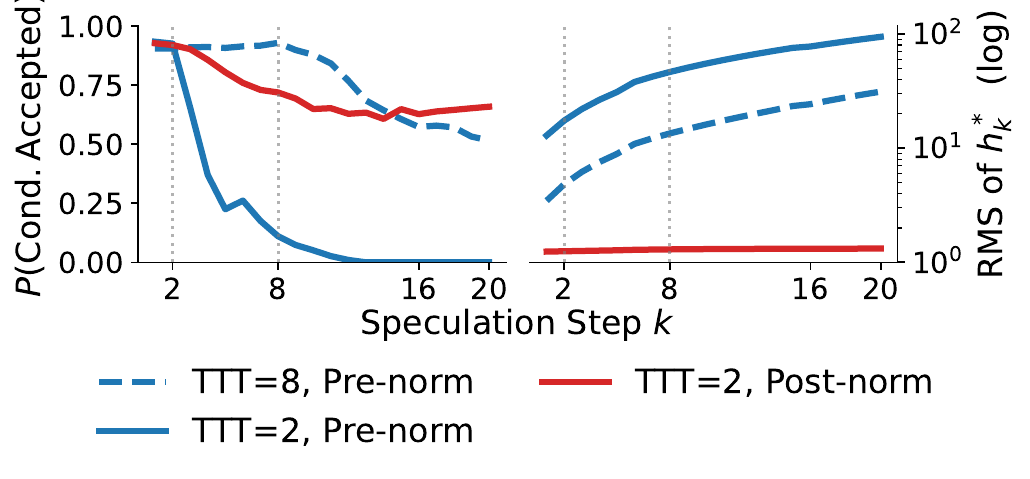}
    \end{subfigure}
    \caption{\normalsize
        Pre-norm vs. Post-norm at various TTT (max training depth). Magnitude and per-step acceptance on coding (HumanEval) and math (GSM8K), $k{=}20$ steps.
    }
    \label{fig:ttt}
    \vspace{-10pt}
\end{wrapfigure}

To test this interpretation, we compare standard pre-norm and the modified post-norm drafters trained with different train-time-test depths. Train-time-test depth denotes the maximum speculation depth used during training. If a drafter learns a depth-invariant autoregressive rule, then training with a small depth should still generalize reasonably to longer speculation chains at inference. In contrast, if the drafter learns depth-specific transformations resembling layers $N, N{+}1, \dots,$ then it should fail beyond the depths observed during training. \Cref{fig:ttt} supports the latter interpretation for pre-norm drafters. A pre-norm drafter trained with train-time-test depth 2 performs well at the first few drafting steps but collapses beyond its training horizon. Its hidden-state magnitude grows rapidly, and its conditional acceptance probability drops sharply at deeper speculation steps.

By contrast, the post-norm drafter trained with the same TTT depth remains stable beyond the training horizon. Applying post-normalization to the drafter hidden state between speculation steps prevents residual-scale accumulation, making the drafter less able to encode depth through magnitude growth. Rather than continuing the verifier as a depth-dependent stack of additional pre-norm transformer layers, the post-norm drafter is regularized toward a stable autoregressive prediction function.

\subsection{Fixing the Attention Sink}
The observation above suggests that hidden-state magnitude accumulation is a major contributor to attention drift. However, attention drift could also be interpreted as a consequence of attention-sink collapse: if the sink weakens during drafting, attention mass must move elsewhere. In this section, we test whether eliminating the sink is sufficient to solve drift as well as the magnitude growth.

To test this, we compare several architectural variants on the same target: a standard pre-norm drafter, a gated-attention drafter, a post-norm drafter, and combining gated-attention and post-norm. The gated-attention variant introduces a learned element-wise gate inside the attention layer (\Cref{ap:gated}), which strongly suppresses reliance on a fixed sink token. \Cref{tab:drafter-variants} summarizes the results.

\begin{table}[h!]
\centering
\caption{Gated attention vs. normalization: chain magnitude of hidden states $h^*_k$, entropy $H$, sink and recent token attention across EAGLE3 drafters for Llama-3.1 8B on MT-Bench (80 prompts over 8 categories and all drafting rounds).}
\begin{tabular}{l ccccc}
\toprule
Drafter            & $h^*_{1\to 8}$ & $H_{1\to 8}$ & sink$_{1\to 8}$ & recent$_{1\to 8}$ \\
\midrule
Pre-norm        & $3.92 \to 14.02$ & $2.04 \to 2.73$ & $0.46 \to 0.08$ & $0.14 \to 0.31$ \\
Gated-Attn                 & $8.41 \to 39.07$ & $2.39 \to 2.13$ & $0.03 \to 0.02$ & $0.29 \to 0.32$  \\
Post-norm              & $1.21 \to 1.20$  & $2.24 \to 2.29$ & $0.11 \to 0.10$ & $0.50 \to 0.53$ \\
Gated $+$ post-norm    & $0.87 \to 0.88$  & $0.63 \to 0.64$ & $0.00 \to 0.00$ & $0.50 \to 0.51$  \\
\bottomrule
\end{tabular}
\label{tab:drafter-variants}
\end{table}

%\textbf{Magnitude and attention are independent failure modes.} Post-norm completely pins the magnitude (\cref{tab:drafter-variants}) and also addresses attention drift, its per-step attention measurements show the \emph{highest} chain mass of any drafter we studied, $0.41$ at $k{=}8$, however the total attention is consistent among prefill and auto-regressive generation and sink is weakened but still there. Gated attention effectively removes the sink and drift ($\leq\!0.03$ throughout), but its chain magnitude grows $5.0\times$ during drafting and starts from a larger $h^*_1 \approx 8.4$. \textbf{Solving attention drift does not eliminate magnitude issue.}

\begin{insight}
\textbf{Magnitude and attention are independent failure modes.} Gated attention nearly eliminates sink attention and removes the visible drift pattern ($\leq\!0.03$ throughout), but its hidden state magnitude grows $5.0\times$ during drafting and starts from a larger $h^*_1 \approx 8.4$. On the other hand,  post-norm pins the hidden-state magnitude across speculation steps and stabilizes the attention pattern over the drafting chain. These results indicate that (i) fixing attention sink eliminates the drift, (ii) attention drift and magnitude accumulation are independent failure modes and that \textbf{solving attention drift does not necessarily eliminate the magnitude issue.}
\end{insight}

\textbf{Role of Entropy.} We also examine whether attention entropy ($H$) explains drafter failures. Each model sits at a different base entropy level, and per-token entropy rises slightly as the speculation depth $k$ grows, the distribution becomes differently-\emph{shaped}, not peakier. However, no consistent relationship between entropy and acceptance length emerges across our pre-norm, post-norm, and gated-attention variants, indicating that drafters learn to compensate for entropy shifts and that entropy alone is not a useful predictor of drafter quality across architectures.

\textbf{Combining the \textit{Gated Attention} with \textit{Post-norm} creates a new pathology.} The Post-norm + Gated variant combines both modifications, and it does what one would hope along each of the two individual axes: chain magnitude is flat, the sink is eliminated, and the recent token attention is stable at $0.50$. But entropy collapses to $H \approx 0.62$ already at pre-fill, about a third of the other drafters' values, corresponding to an effective attention support of roughly two positions ($e^{0.62} \approx 1.85$). Applying both changes appears to over-regularize the drafter: the attention distribution collapses onto roughly two tokens.

\FloatBarrier

\subsection{Noise and Error Accumulation}

The drafter consumes its own predictions as inputs to subsequent speculation steps, and these self-generated hidden states are noisier than the clean verifier states seen at $t{=}1$. The $\alpha$-noise sweep therefore tests how gracefully each architecture absorbs imprecise hidden-state inputs, determining how quickly errors compound with depth. \textbf{Post-norm tolerates an order of magnitude more perturbation than pre-norm} on the hidden pathway (58\% vs.\ 5\% at $\alpha{=}0.5$), offering a mechanistic explanation for its better deep-chain behavior: it accumulates less error per speculation step. We further hypothesize that this tolerance may translate to robustness under other small hidden-state perturbations, such as those induced by verifier quantization or mild distribution shift.

\begin{table}[h!]
\centering
\caption{Drafter pathway reliance under noise injection. Each cell reports acceptance length as a percentage of the model's no-perturbation baseline (parenthesized in the model column). Noise is scaled per-tensor RMS: $x \leftarrow x + \alpha\cdot\mathrm{rms}(x)\cdot\varepsilon$.}
\label{tab:pathway_reliance}
\begin{tabular}{ll cccccc}
\toprule
Model & Target & $\alpha{=}0.1$ & $\alpha{=}0.25$ & $\alpha{=}0.5$ & $\alpha{=}1.0$ & $\alpha{=}2.0$ & $\alpha{=}\infty$ \\
\midrule
\multirow{2}{*}{Pre-norm (3.06)}
              & Hidden States  & \phantom{0}82\% & \phantom{0}28\% & \phantom{0}5\% & \phantom{0}0\% & \phantom{0}0\% & \phantom{0}0\% \\
              & Embeddings   & \phantom{0}99\% & \phantom{0}93\% & 86\%           & 64\%           & 27\%           & \phantom{0}9\% \\
\addlinespace
\multirow{2}{*}{Post-norm (3.16)}
              & Hidden States  & \phantom{0}99\% & \phantom{0}86\% & 58\%           & 22\%           & \phantom{0}5\% & \phantom{0}2\% \\
              & Embeddings   & \phantom{0}98\% & 100\%           & 93\%           & 75\%           & 38\%           & \phantom{0}9\% \\
\bottomrule
\end{tabular}
\end{table}

\textbf{Does artificially shrinking the magnitude fix attention drift?} We observed that fixing the drift does not necessarily fix the magnitude accumulation. However we are not sure whether fixing the magnitude alone would fix attention drift. We have created an inference-time controlled experiment, the drafter's hidden states ($h_{out}$) were normalized to match the \textit{fc}'s magnitude during inference. \Cref{tab:magnitude_pin} shows that normalizing the magnitude without matched training hurts the accuracy in both cases, pre-norm being effected even more than post-norm. However we observe one interesting thing, the attention drift is significantly lessened on pre-norm architecture, while still happening weakly. This shows that magnitude accumulation is one of the contributors to the attention drift but not the only one. Note that the experiment isolates the magnitude-attention link at inference but does not establish what a magnitude-controlled drafter would learn end-to-end.

\begin{table}[h!]
\centering
\caption{Effect of pinning the hidden-state RMS to the FC's RMS on Llama 3.1 8B during test time.}
\label{tab:magnitude_pin}
\begin{tabular}{ll rr cc r}
\toprule
Model     & Mode      & Accept & $\Delta$Acc.\ & Sink $k{=}0$ & Sink $k{=}7$ & Chain $k{=}7$ \\
\midrule
Pre-norm        & Regular  & 3.06   & --            & 0.40         & \textbf{0.14}         & 0.31 \\
Pre-norm        & Normed    & 1.33   & $-56\%$       & 0.41         & 0.29         & 0.24 \\
\addlinespace
Post-norm & Regular  & 3.15   & --            & 0.11         & 0.11         & 0.35 \\
Post-norm & Normed    & 2.09   & $-34\%$       & 0.11         & 0.13         & 0.33 \\
\bottomrule
\end{tabular}
\end{table}

\textbf{Role of Training Window.} A possible additional contributor to attention drift is the training procedure itself. Most EAGLE-style trainers use a fixed context window during train-time testing (TTT), with the oldest tokens dropped from the window as drafting proceeds. This means the drafter is trained on inputs where early prompt positions, including the sink token, are progressively pushed out of context as speculation depth increases. The drafter may therefore learn to reduce reliance on these positions over the chain. While our main results identify hidden-state magnitude accumulation as a key contributor to drift, this training-window effect may further amplify sink weakening. We leave a detailed study of how training-window construction interacts with attention drift to future work.

\subsection{Attention Drift in Other Architectures}

\begin{figure}[h]
    \centering
    \includegraphics[width=\linewidth]{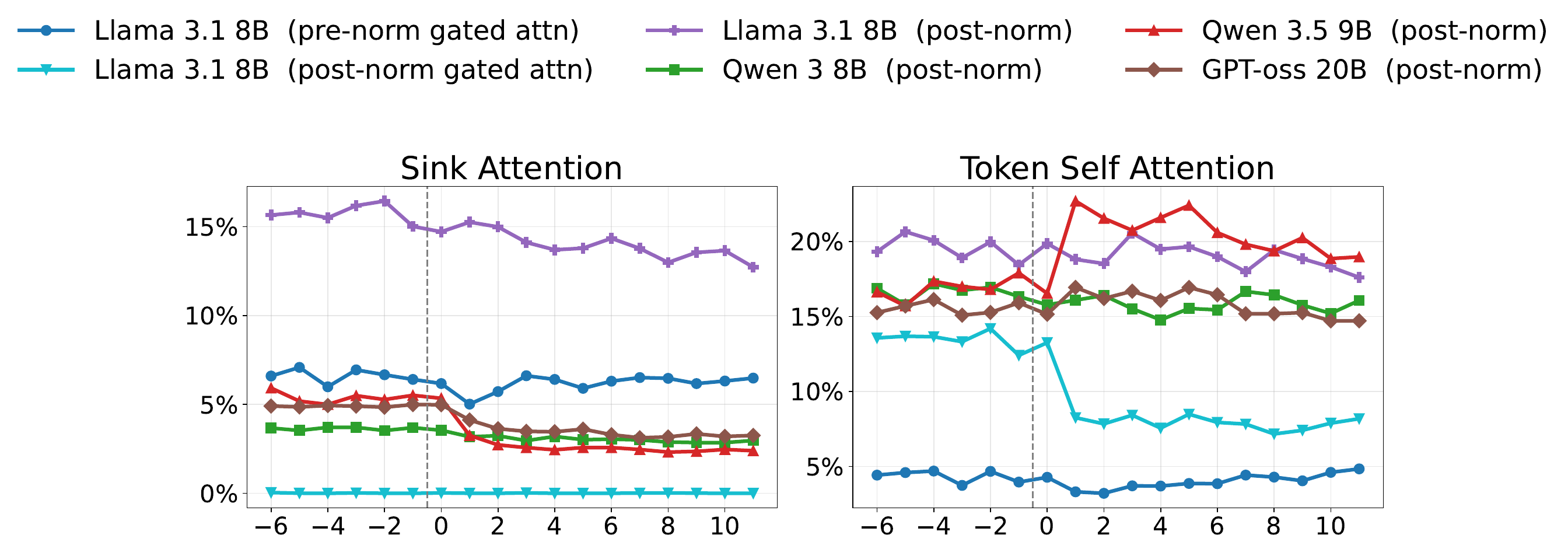}
    \caption{Attention Sink and Self-Token Attention for each model that fixes the attention drift.}
    \label{fig:all-drift-fixes}
\end{figure}

Sections 4.2 and 4.3 propose two fixes for EAGLE: post-norm and gated attention. However our experiments so far cover only one model family. Here we examine whether these fixes generalize. \Cref{fig:all-drift-fixes} shows that across drafter–target pairs, both architectural changes stabilize sink attention and self-token attention. The gated-attention models (Qwen 3.5 and the post-norm gated variant of Llama 3.1) settle at different absolute attention levels than the post-norm-only models, but they similarly do not exhibit progressive drift across the drafting chain.

\textbf{MTP Heads.} We observe a different drift profile in MTP heads compared to pre-norm EAGLE-3: drift onsets sharply within the first few speculation steps and then stabilizes at a new attention baseline, rather than progressing gradually across the entire chain (\Cref{fig:mtpattn}). Sink attention drops sharply at speculation start, and attention to the most recently drafted token rises and then plateaus after a few steps. This is notable because MTP uses a post-norm architecture, where we expected drift to be largely suppressed. We see a similar sharp-then-stabilize pattern in our gated-attention variants (\Cref{fig:all-drift-fixes}), suggesting that post-norm combined with attention reweighting produces a distinct dynamic from gradual pre-norm drift.

\begin{figure}[h!]
    \centering
    \includegraphics[width=0.48\linewidth]{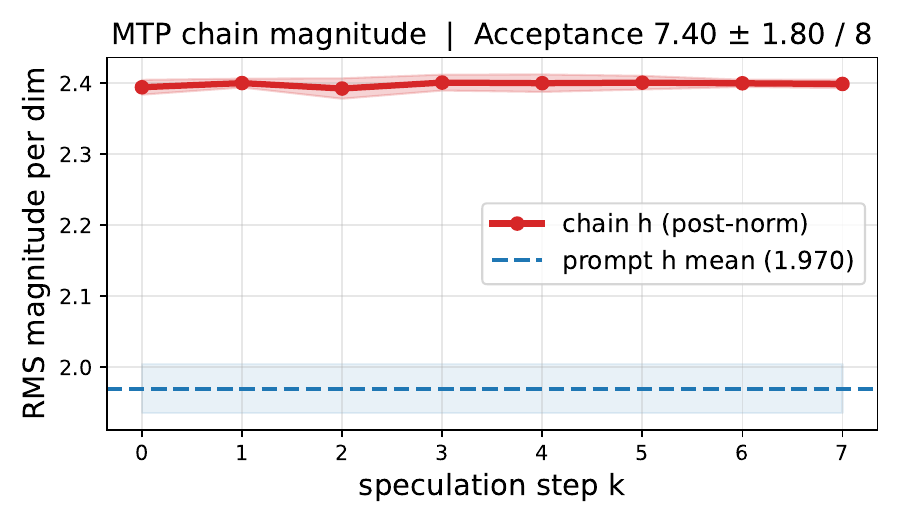}
    \caption{Magnitude drift on Qwen3.5 9B MTP heads}
    \label{fig:mtpgrowth}
\end{figure}

This is also consistent with our finding that \textbf{magnitude accumulation is one contributor to drift but not the only mechanism}. We focus our diagnosis and intervention on EAGLE-3, which trains a separate drafter with its own LM head post-hoc against the frozen target's output distribution. MTP, in contrast, reuses the target's LM head and is trained jointly with the target during pretraining, which may contribute to the observed effect. The exact training procedure for Qwen3.5's MTP heads has not been publicly disclosed; characterizing the role of training loss versus architecture in the emergence of attention drift is left as future work.

\section{Performance Impact}

In this section, we evaluate different solutions to the attention and magnitude drift problem. We report acceptance length excluding the bonus token except for SGLang benchmarks, to compare raw performance across speculative decoding models. We trained drafters for four target models spanning different architectures: Llama 3.1 8B (dense), Qwen 3 8B (dense thinking), Qwen 3.5 9B (GDN-hybrid thinking), and GPT-OSS 20B (sparse MoE thinking).

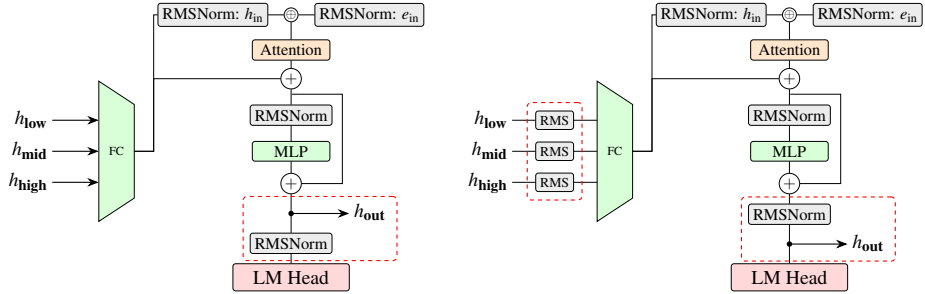
\begin{figure}[h!]
    \centering
    \resizebox{0.8\linewidth}{!}{\begin{tikzpicture}[
    scale=1.0, every node/.style={scale=1.0},
    block/.style={draw, rounded corners=2pt, minimum width=3cm, minimum height=0.7cm, font=\Large},
    smallblock/.style={draw, rounded corners=2pt, minimum width=2cm, minimum height=0.5cm, font=\large},
    tinynorm/.style={draw, rounded corners=2pt, minimum width=0.8cm, minimum height=0.45cm, fill=gray!15, font=\normalsize},
    tok/.style={font=\Large\bfseries, text height=1.5ex, text depth=0.25ex},
    arr/.style={-{Stealth[length=3mm]}, thick},
    res/.style={thick, gray, rounded corners=3pt},
]

% =========== LEFT (Original) ===========

\node[smallblock, fill=gray!15] (norm-h) {RMSNorm: $h_{\text{in}}$};
\node[smallblock, fill=gray!15, right=1.2cm of norm-h] (norm-e) {RMSNorm: $e_{\text{in}}$};

\node[draw, circle, inner sep=0pt, font=\normalsize] at ($(norm-h)!0.5!(norm-e)$) (concat) {$\oplus$};
\draw (norm-h) -- (concat);
\draw (norm-e) -- (concat);

\node[smallblock, fill=orange!20, below=0.45cm of concat] (attn) {Attention};
\draw[thin] (concat) -- (attn);

\node[draw, circle, inner sep=1pt, font=\large, below=0.2cm of attn] (add1) {$+$};
\draw[thin] (attn) -- (add1);

\node[smallblock, fill=gray!15, below=0.4cm of add1] (norm2) {RMSNorm};
\draw[thin] (add1) -- (norm2);

\node[smallblock, fill=green!15, below=0.4cm of norm2] (mlp) {MLP};
\draw[thin] (norm2) -- (mlp);

\node[draw, circle, inner sep=1pt, font=\large, below=0.3cm of mlp] (add2) {$+$};
\draw[thin] (mlp) -- (add2);

\draw[thin] ([yshift=-4mm,xshift=-2.5mm]add1.east) -- ++(1.3cm,0) |- (add2.east);

\node[smallblock, fill=gray!15, below=1.0cm of add2] (norm-out) {RMSNorm};
\draw[thin] (add2) -- (norm-out);

\node[block, fill=red!15, below=0.25cm of norm-out] (lmhead) {LM Head};
\draw[thin] (norm-out) -- (lmhead);

\coordinate (h-tap) at ($(add2.south)!0.5!(norm-out.north)$);
\node[tok, right=1.5cm of h-tap] (h-out) {$h_{\text{out}}$};
\draw[arr] (h-tap) -- (h-out);
\fill (h-tap) circle (2pt);
\node[draw=red, dashed, thick, rounded corners=3pt, inner sep=5pt,
      fit=(h-tap)(h-out)(norm-out), scale=1.0] {};

% --- FC + hidden states on the LEFT, at attn's vertical level ---
\node[minimum width=0.9cm, minimum height=2cm] at ([xshift=-4.5cm]mlp) (fc) {};
\draw[fill=green!15]
    ([yshift=8mm]fc.north west) -- ([yshift=2mm]fc.north east) --
    ([yshift=-2mm]fc.south east) -- ([yshift=-8mm]fc.south west) -- cycle;
\node[font=\small] at (fc) {FC};

\node[tok, left=1.2cm of fc, yshift= 8mm] (hlow)  {$h_{\text{low}}$};
\node[tok, left=1.2cm of fc]               (hmid)  {$h_{\text{mid}}$};
\node[tok, left=1.2cm of fc, yshift=-8mm] (hhigh) {$h_{\text{high}}$};
\draw[arr] (hlow.east)  -- (fc.west |- hlow);
\draw[arr] (hmid.east)  -- (fc.west |- hmid);
\draw[arr] (hhigh.east) -- (fc.west |- hhigh);

% FC output routes up to h-in
\draw[thin] (fc.east) -- ++(0.5cm,0) |- (norm-h.west);

% Residual
\draw[thin] (fc.east) -- ++(0.5cm,0) |- (add1.west);

% =========== RIGHT (Ours) ===========

\node[smallblock, fill=gray!15, right=10cm of norm-h] (norm-h2) {RMSNorm: $h_{\text{in}}$};
\node[smallblock, fill=gray!15, right=1.2cm of norm-h2] (norm-e2) {RMSNorm: $e_{\text{in}}$};

\node[draw, circle, inner sep=0pt, font=\normalsize] at ($(norm-h2)!0.5!(norm-e2)$) (concat2) {$\oplus$};
\draw (norm-h2) -- (concat2);
\draw (norm-e2) -- (concat2);

\node[smallblock, fill=orange!20, below=0.45cm of concat2] (attn2) {Attention};
\draw[thin] (concat2) -- (attn2);

\node[draw, circle, inner sep=1pt, font=\large, below=0.2cm of attn2] (add1b) {$+$};
\draw[thin] (attn2) -- (add1b);

\node[smallblock, fill=gray!15, below=0.4cm of add1b] (norm2b) {RMSNorm};
\draw[thin] (add1b) -- (norm2b);

\node[smallblock, fill=green!15, below=0.4cm of norm2b] (mlp2) {MLP};
\draw[thin] (norm2b) -- (mlp2);

\node[draw, circle, inner sep=1pt, font=\large, below=0.3cm of mlp2] (add2b) {$+$};
\draw[thin] (mlp2) -- (add2b);

\draw[thin] ([yshift=-4mm,xshift=-2.5mm]add1b.east) -- ++(1.3cm,0) |- (add2b.east);

\node[smallblock, fill=gray!15, below=0.25cm of add2b] (norm-out2) {RMSNorm};
\draw[thin] (add2b) -- (norm-out2);

\node[block, fill=red!15, below=1.0cm of norm-out2] (lmhead2) {LM Head};
\draw[thin] (norm-out2) -- (lmhead2);

\coordinate (h-tap2) at ($(norm-out2.south)!0.5!(lmhead2.north)$);
\node[tok, right=1.5cm of h-tap2] (h-out2) {$h_{\text{out}}$};
\draw[arr] (h-tap2) -- (h-out2);
\fill (h-tap2) circle (2pt);
\node[draw=red, dashed, thick, rounded corners=3pt, inner sep=5pt,
      fit=(h-tap2)(h-out2)(norm-out2), scale=1.0] {};

% --- FC + RMSNorms + hidden states on the LEFT, at mlp's vertical level ---
\node[minimum width=0.9cm, minimum height=2cm] at ([xshift=-4.5cm]mlp2) (fc2) {};
\draw[fill=green!15]
    ([yshift=8mm]fc2.north west) -- ([yshift=2mm]fc2.north east) --
    ([yshift=-2mm]fc2.south east) -- ([yshift=-8mm]fc2.south west) -- cycle;
\node[font=\small] at (fc2) {FC};

\node[tinynorm, left=0.6cm of fc2, yshift= 8mm] (n-low)  {RMS};
\node[tinynorm, left=0.6cm of fc2]              (n-mid)  {RMS};
\node[tinynorm, left=0.6cm of fc2, yshift=-8mm] (n-high) {RMS};

\node[tok, left=0.6cm of n-low]  (hl2) {$h_{\text{low}}$};
\node[tok, left=0.6cm of n-mid]  (hm2) {$h_{\text{mid}}$};
\node[tok, left=0.6cm of n-high] (hh2) {$h_{\text{high}}$};

\draw[thin] (hl2) -- (n-low);
\draw[thin] (hm2) -- (n-mid);
\draw[thin] (hh2) -- (n-high);

\draw[thin] (n-low.east)  -- (fc2.west |- n-low);
\draw[thin] (n-mid.east)  -- (fc2.west |- n-mid);
\draw[thin] (n-high.east) -- (fc2.west |- n-high);

% Red bounding box around the new RMSNorms only
\node[draw=red, dashed, thick, rounded corners=3pt, inner sep=6pt,
      fit=(n-low)(n-mid)(n-high)] {};

% FC output routes up to h-in (norm-h2)
\draw[thin] (fc2.east) -- ++(0.5cm,0) |- (norm-h2.west);

% Residual from FC output to add1b
\draw[thin] (fc2.east) -- ++(0.5cm,0) |- (add1b.west);

\end{tikzpicture}}
    \caption{Standard \emph{Pre-norm} (Left) vs proposed \emph{Post-norm} (Right) drafter architectures.}
    \label{fig:postnorm-arch}
\end{figure}

The proposed post-norm architecture places individual RMSNorms after each target hidden state $h_{low},h_{mid},h_{high}$ and accumulates the drafter's hidden states \emph{after} the RMSNorm (\Cref{fig:postnorm-arch}). By using post-norm, we were able to reduce the TTT length from 8 to 4, cutting training time by roughly one third without impacting performance. We also trained baseline models using the same training data and configuration with the regular pre-norm architecture for a fair comparison. \textbf{Post-norm improved acceptance length consistently across all four models, with gains ranging up to $\mathbf{12\%}$} (\Cref{fig:placeholder}, full per-model results in Appendix~\ref{ap:results}).

\begin{figure}[h!]
    \centering
    \includegraphics[width=1.0\linewidth]{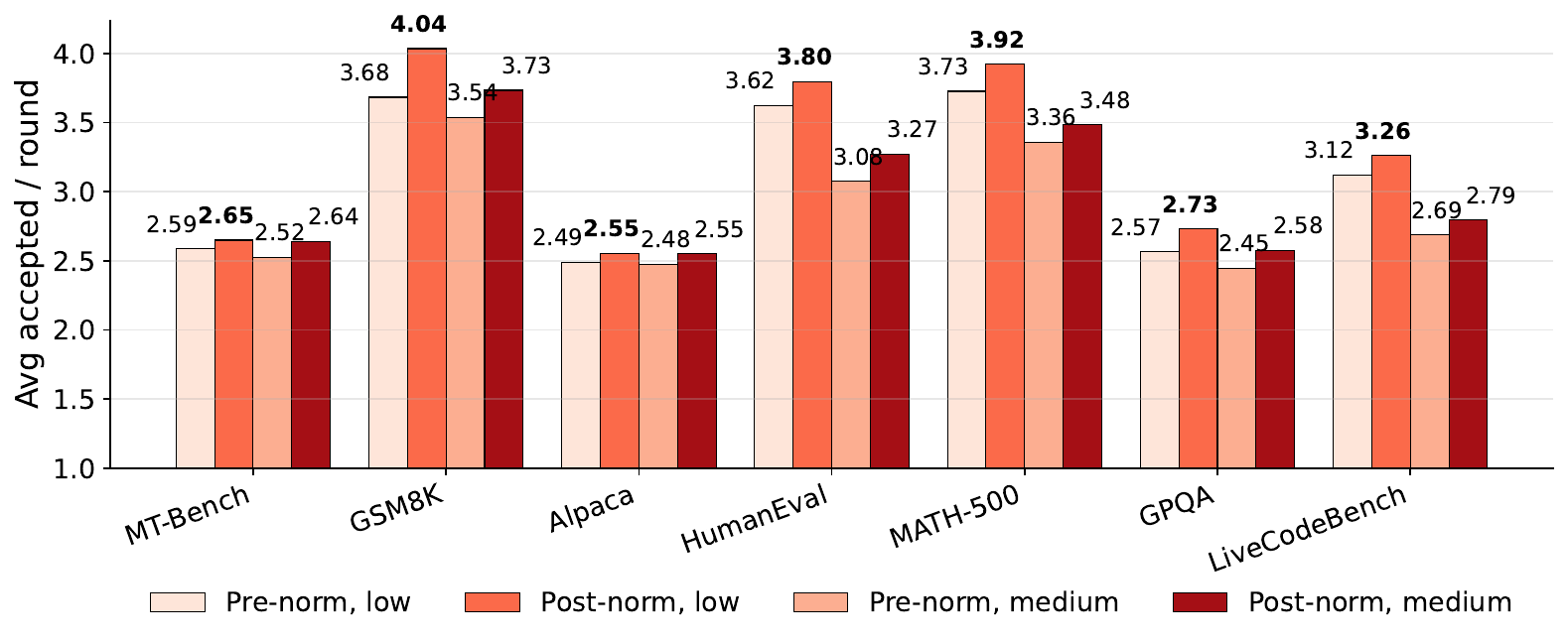}
    \caption{GPT-OSS 20B results on SGLang, acceptance length includes bonus token, temp $=0.7$. Proposed post-norm shows consistent improvements over standard pre-norm.}
    \label{fig:placeholder}
\end{figure}

\subsection{Template Sensitivity}

Performance of speculative decoding models varies heavily across different deployment settings, system prompts, and chat templates. Since drafters are not trained on raw pre-training data but on processed supervised fine-tuning (SFT) conversations, we observed they heavily overfit to certain elements of the chat template. Templates may be perturbed intentionally (e.g., disabling reasoning tags to save tokens) or unintentionally, as reported in inference engines and benchmarks. We test how well our post-norm architecture generalizes beyond the chat format and improves resilience to deployment conditions.

\begin{figure}[h!]
    \centering
    \includegraphics[width=0.55\linewidth]{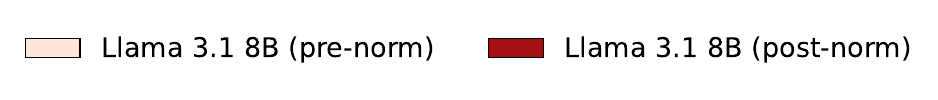}
    \begin{minipage}{0.48\linewidth}
        \centering
        \includegraphics[width=\linewidth]{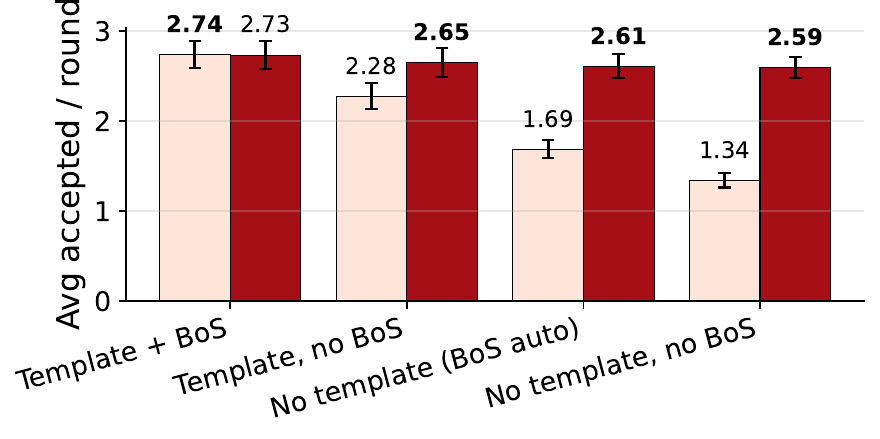}
        \caption{Template sensitivity of Llama 3.1 8B drafter pre-norm and post-norm. Temp = 0.7.}
        \label{fig:template}
    \end{minipage}
    \hfill
    \begin{minipage}{0.48\linewidth}
        \centering
        \includegraphics[width=\linewidth]{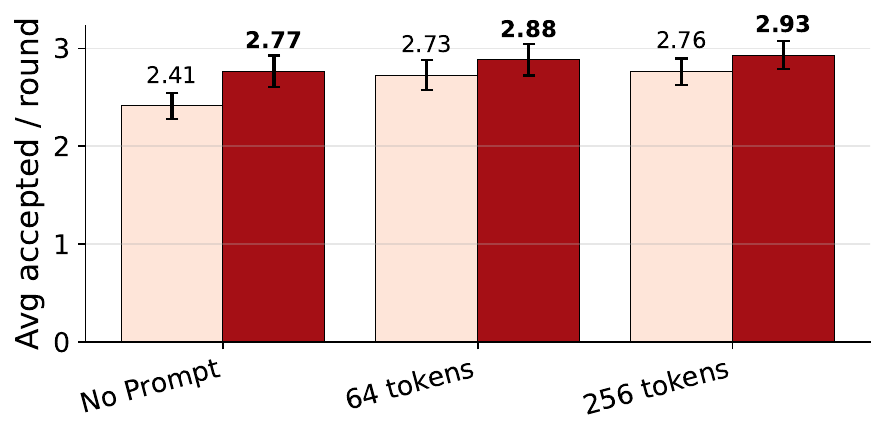}
        \caption{System prompt length's effect on accuracy for Llama 3.1 8B.}
        \label{fig:sysprompt}
    \end{minipage}
\end{figure}

We designed two experiments to test how well our post-norm architecture generalizes beyond the training-time chat format and prompt length. In the first, we removed the chat template and sink token to measure sensitivity (\Cref{ap:template}). \textbf{Post-norm was significantly more resilient to every kind of disruption}, losing at most 5\% accuracy in the worst case, whereas pre-norm dropped 52\% (\Cref{fig:template}). Temperature set to $0.7$ to prevent token-level repetition we frequently observed under template perturbation.

\begin{figure}[h!]
    \centering
    \includegraphics[width=1.0\linewidth]{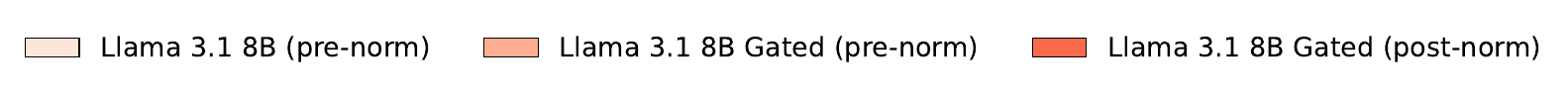}
    \begin{minipage}{0.48\linewidth}
        \centering
        \includegraphics[width=\linewidth]{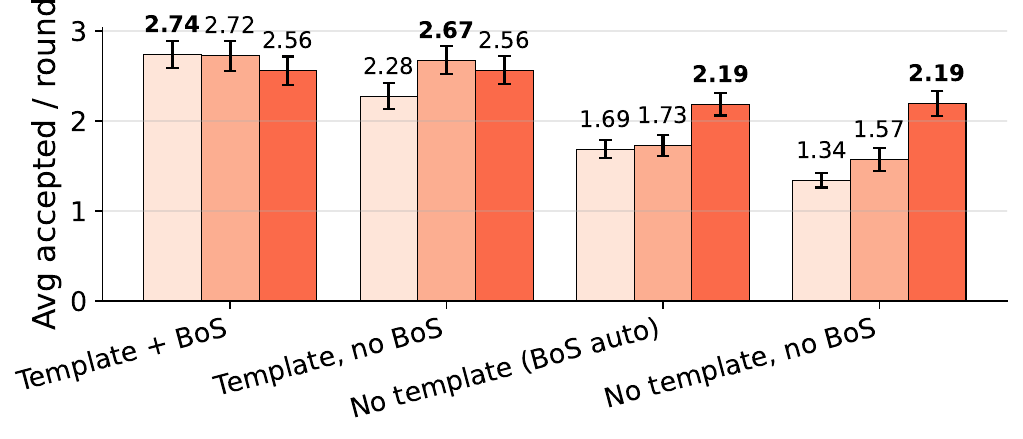}
        \caption{Gating is not sensitive to BoS, but it still overfits to other tokens in the template.}
        \label{fig:gated-bos}
    \end{minipage}
    \hfill
    \begin{minipage}{0.48\linewidth}
        \centering
        \includegraphics[width=\linewidth]{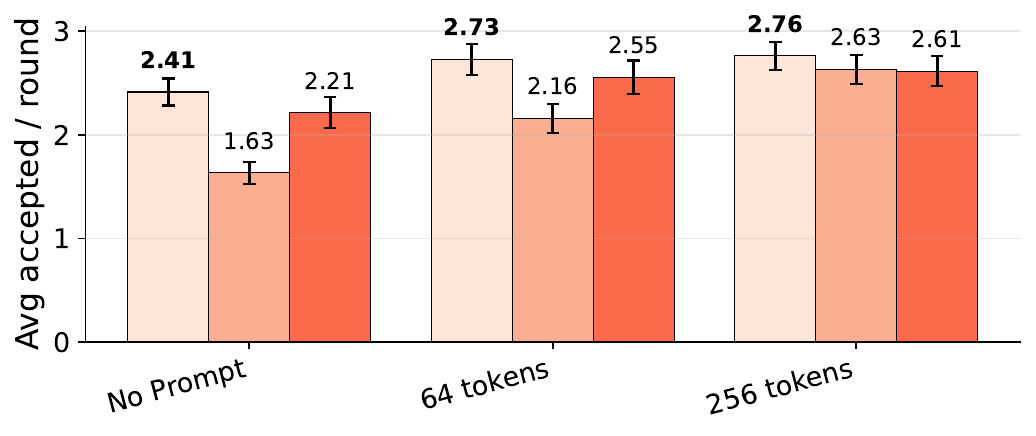}
        \caption{Gated attention overfits to the system prompt, hurting accuracy badly.}
        \label{fig:gated-overfit}
    \end{minipage}
\end{figure}

\begin{insight}
\textbf{Is template sensitivity an artifact of attention sinks?} To test this, we compared post-norm against a Gated Attention model, which eliminates the attention sink and, thus, the sensitivity to the beginning-of-sequence (BoS) token. The gated model(s) still showed high template sensitivity and in fact developed \emph{higher} sensitivity to system prompt length, with lower overall accuracy (\Cref{fig:gated-bos,fig:gated-overfit}). This shows that \textbf{removing the sink alone is not enough to reduce template fragility} and that fixing hidden-residual dynamics is required. We observe similar template sensitivity patterns on Qwen3-8B and GPT-oss 20B (\Cref{tab:exp1-qwen,tab:exp1-gptoss}), where post-norm provides 5--35\% gains under template perturbation, with the gap widening as templates become more disrupted.
\end{insight}

\begin{table}[h!]
\centering
\begin{minipage}[t]{0.48\textwidth}
\centering
\small
\caption{\textbf{Qwen3-8B}: Avg accepted draft tokens per verification round under different prompt-template manipulations on MT-Bench.}
\label{tab:exp1-qwen}
\begin{tabular}{l r r}
\toprule
Condition & Pre-norm & Post-norm \\
\midrule
template + BoS & 1.80 $\pm$ 0.07 & \textbf{1.90} $\pm$ 0.07 \\
template, no BoS & 1.76 $\pm$ 0.06 & \textbf{1.87} $\pm$ 0.06 \\
no template (BoS auto) & 1.71 $\pm$ 0.11 & \textbf{2.15} $\pm$ 0.12 \\
no template, no BoS & 1.64 $\pm$ 0.09 & \textbf{2.21} $\pm$ 0.10 \\
\bottomrule
\end{tabular}
\end{minipage}%
\hfill
\begin{minipage}[t]{0.48\textwidth}
\centering
\small
\caption{\textbf{GPT-oss 20B}: Avg accepted draft tokens per verification round under different prompt-template manipulations on MT-Bench.}
\label{tab:exp1-gptoss}
\begin{tabular}{l r r}
\toprule
Condition & Pre-norm & Post-norm \\
\midrule
template + BoS & 1.66 $\pm$ 0.13 & \textbf{1.80} $\pm$ 0.14 \\
template, no BoS & 1.55 $\pm$ 0.11 & \textbf{1.73} $\pm$ 0.13 \\
no template (BoS auto) & 1.43 $\pm$ 0.07 & \textbf{1.75} $\pm$ 0.07 \\
no template, no BoS & 1.29 $\pm$ 0.05 & \textbf{1.65} $\pm$ 0.07 \\
\bottomrule
\end{tabular}
\end{minipage}
\end{table}

Another factor in template sensitivity may lie in how the loss is constructed during training. In the EAGLE series, the loss is computed only on assistant tokens; user tokens don't contribute to the loss directly. So while parameters are shared and trained on the full sequence, the model's outputs at user positions are not directly constrained, only their utility for downstream assistant-position predictions is. This asymmetry could show up especially when the chat template is changed and the boundary between supervised and unsupervised positions change from where the model was trained to expect it.

\textbf{System Prompt Length.} In our second experiment with temperature 0, we varied system prompt length (Llama's default prompt, trimmed to different lengths). Post-norm beat pre-norm at every length tested (\Cref{fig:sysprompt}). More importantly, \textbf{post-norm was substantially more robust to system prompt length variation}: post-norm degraded at most 5\% from its peak, while pre-norm dropped 13\%. Post-norm's worst case still beats pre-norm's best.

The cross-model differences in this sensitivity correlate with training-time prompt distribution. Llama's default system prompt is around 500 characters, while Qwen3 and GPT-OSS are trained with much shorter system prompts (30--40 characters). This may explain why Qwen3 and GPT-OSS are more resilient to system-prompt length changes overall, and why they perform best at prompts around 64 tokens (\Cref{tab:exp2-qwen,tab:exp2-gptoss}).

\begin{table}[h!]
\centering
\begin{minipage}[t]{0.48\textwidth}
\centering
\small
\caption{\textbf{Qwen3-8B}: Avg accepted draft tokens per verification round as the system context grows from 0 to 256 tokens.}
\label{tab:exp2-qwen}
\begin{tabular}{l r r}
\toprule
Condition & Pre-norm & Post-norm \\
\midrule
no system & 1.82 $\pm$ 0.06 & \textbf{1.94} $\pm$ 0.06 \\
system 64 tok & 1.92 $\pm$ 0.06 & \textbf{2.00} $\pm$ 0.07 \\
system 256 tok & 1.89 $\pm$ 0.07 & \textbf{1.96} $\pm$ 0.07 \\
\bottomrule
\end{tabular}
\end{minipage}%
\hfill
\begin{minipage}[t]{0.48\textwidth}
\centering
\small
\caption{\textbf{GPT-oss 20B}: Avg accepted draft tokens per verification round as the prepended system context grows from 0 to 256 tokens.}
\label{tab:exp2-gptoss}
\begin{tabular}{l r r}
\toprule
Condition & Pre-norm & Post-norm \\
\midrule
no system & 1.86 $\pm$ 0.13 & \textbf{2.01} $\pm$ 0.14 \\
system 64 tok & 1.94 $\pm$ 0.12 & \textbf{2.06} $\pm$ 0.13 \\
system 256 tok & 1.85 $\pm$ 0.12 & \textbf{2.02} $\pm$ 0.14 \\
\bottomrule
\end{tabular}
\end{minipage}
\end{table}

\subsection{Long Context}

Another challenge for speculative decoding models is long context. Drafters are relatively cheap to train and are usually trained with a short context window such as 4096 tokens, which makes them easier and more affordable to train. However, they fail catastrophically outside their trained context length. LLMs have developed various techniques to handle this, one being sliding window attention (SWA), where the model's effective context is a fixed window over the most recent tokens. This makes intuitive sense for drafters in particular: they are weaker by design, and their predictions on easy tokens shouldn't depend on long-range context. SWA has been studied for long-context speculative decoding from both serving-systems \citep{magicdec} and drafter-training angles \citep{longspec}. We use SWA as a long-context evaluation tool rather than as a method we propose; our post-norm fix is orthogonal to these works and can be combined with existing long-context speculative decoding techniques.

\begin{figure}[h!]
    \centering
    \includegraphics[width=0.55\linewidth]{images/legend_exp1.pdf}
    \begin{minipage}{0.48\linewidth}
        \centering
        \includegraphics[width=\linewidth]{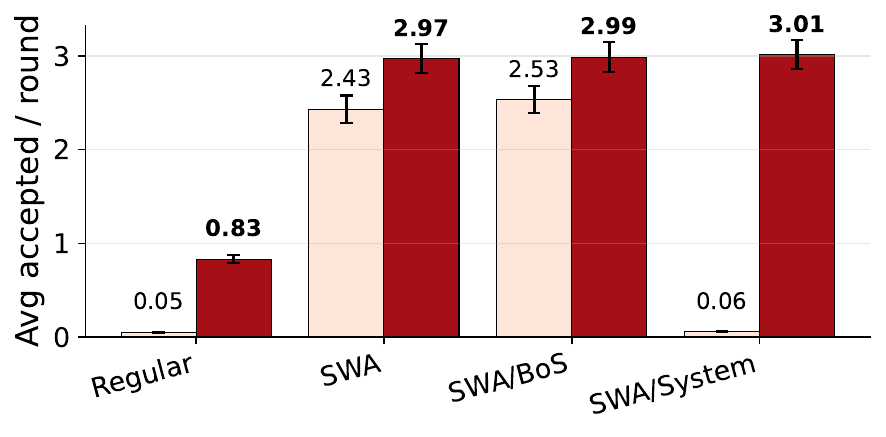}
        \caption{Accuracy of Llama 3.1 8B drafter on various long-context multi-turn chat settings.}
        \label{fig:multiturn}
    \end{minipage}
    \hfill
    \begin{minipage}{0.48\linewidth}
        \centering
        \includegraphics[width=\linewidth]{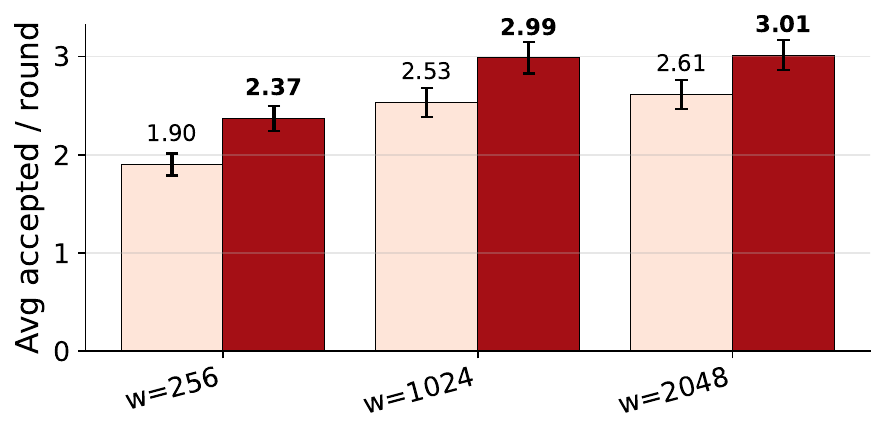}
        \caption{Effect of window size on prediction accuracy.}
        \label{fig:windowsize}
    \end{minipage}
\end{figure}

\begin{figure}[b!]
    \resizebox{1.0\linewidth}{!}{\begin{tikzpicture}[
    tok/.style={minimum width=0.6cm, minimum height=0.6cm, draw, font=\tiny, inner sep=0pt},
]

% === SWA ===
\node[font=\Large\bfseries] at (3.9, 0.8) {SWA};

\foreach \i in {0,...,12} {
    \ifnum\i>8
        \node[tok, fill=orange!30] (a-\i) at (\i*0.65, 0) {};
    \else
        \node[tok, fill=gray!10] (a-\i) at (\i*0.65, 0) {};
    \fi
}

% === SWA + BoS ===
\begin{scope}[xshift=10cm]
\node[font=\Large\bfseries] at (3.9, 0.8) {SWA + BoS};

\foreach \i in {0,...,12} {
    \ifnum\i=0
        \node[tok, fill=orange!30, 
            postaction={pattern=north east lines, pattern color=orange!80},
            postaction={pattern=north west lines, pattern color=orange!80}
        ] (b-\i) at (\i*0.65, 0) {};
    \else
        \ifnum\i>8
            \node[tok, fill=orange!30] (b-\i) at (\i*0.65, 0) {};
        \else
            \node[tok, fill=gray!10] (b-\i) at (\i*0.65, 0) {};
        \fi
    \fi
}
\end{scope}

% === SWA + System Prompt ===
\begin{scope}[xshift=20cm]
\node[font=\Large\bfseries] at (3.9, 0.8) {SWA + System Prompt};

\foreach \i in {0,...,12} {
    \ifnum\i<3
        \node[tok, fill=orange!30,
                postaction={pattern=north east lines, pattern color=orange!80},
                postaction={pattern=north west lines, pattern color=orange!80}
            ] (c-\i) at (\i*0.65, 0) {};
    \else
        \ifnum\i>8
            \node[tok, fill=orange!30] (c-\i) at (\i*0.65, 0) {};
        \else
            \node[tok, fill=gray!10] (c-\i) at (\i*0.65, 0) {};
        \fi
    \fi
}
\end{scope}

\end{tikzpicture}}
    \caption{Attended tokens (marked in orange) in different SWA implementations.}
    \label{fig:swa}
\end{figure}
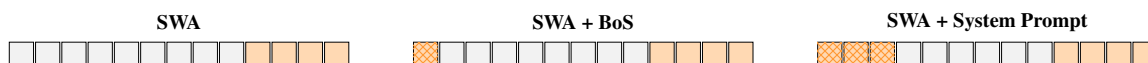

We created a multi-turn conversation benchmark with context length beyond our models' trained length. A multi-turn setup isolates the model's long-context capability from degradation caused by unrelated filler text (results summarized in \Cref{fig:swa}). Under \textbf{full attention} (regular), the pre-norm model fails catastrophically, dropping to $0.05$ average acceptance length, while post-norm drops to $0.83$, also unusable, but $15\times$ better. With \textbf{SWA}, most of the accuracy is rescued: pre-norm recovers $88\%$ of its single-turn baseline, while post-norm matches and slightly exceeds its single-turn performance. Since pre-norm develops a strong sink, \textbf{carrying the BoS token} (SWA/BoS) through the window helps it more (up to $91\%$ recovery), with only a minor bump for post-norm. Following on this, we hypothesized that \textbf{carrying the system prompt} (SWA/System) would add additional context. For post-norm this added another $1\%$, but pre-norm collapsed to near-zero, it could not accommodate the wider range of positional embeddings introduced by the longer prefix. \textbf{Across all SWA conditions, post-norm outperformed pre-norm by $\mathbf{20\%}$.}

\textbf{True long-context performance.} In addition to our multi-turn benchmark, we evaluate on \textit{LongBench}~\cite{longbench} across three task categories: summarization, few-shot learning, and coding. Absolute task scores are lower than in the multi-turn setting due to the inherent difficulty of long-context understanding. \textbf{Post-norm outperformed pre-norm by $\mathbf{20}$--$\mathbf{25\%}$ across all categories and context lengths (\Cref{tab:longbench})}. On the government report summarization task, adding the system prompt yielded an additional $8\%$ gain, suggesting that informative prefixes can further help SWA-based drafters recover context, though we leave a broader study of this effect to future work.

\begin{table}[h!]
\centering
\small
\caption{LongBench: avg accepted draft tokens per round, per task and SWA mode. Window=1024.}
\begin{tabular}{l| rr | rr}
\toprule
Task & \multicolumn{2}{c}{SWA + BoS} & \multicolumn{2}{c}{SWA + Sys Prompt} \\
 & Pre-norm & Post-norm & Pre-norm & Post-norm \\
\midrule
\texttt{Gov Report (Summarization)} & 1.35 & 1.57 & 0.47 & \textbf{1.70} \\
\texttt{Samsum (Few-Shot)} & 1.04 & \textbf{1.31} & 0.39 & \textbf{1.31} \\
\texttt{Repobench (Coding)} & 1.85 & \textbf{2.26} & 0.26 & 2.25 \\
\bottomrule
\end{tabular}
\label{tab:longbench}
\end{table}

\textbf{How much does the drafter rely on long-range context vs. recent tokens?} To answer this, we varied the SWA window size (\Cref{fig:windowsize}). Across all three window sizes tested, post-norm maintained a consistent gap over pre-norm. Even a small 256-token window recovers $80\%$ of the full-context baseline; performance saturates around 1024 tokens, with diminishing returns beyond that. This suggests \textbf{the model mostly cares about recent tokens and SWA can substantially reduce compute without meaningfully hurting drafter accuracy.}

\section{Related Work}

\textbf{Speculative decoding} accelerates LLM inference by drafting candidate tokens and verifying them in parallel \citep{leviathan2023fast}. Several drafter designs have been proposed: Medusa \citep{cai2024medusa} attaches multiple prediction heads to the target, Hydra \citep{ankner2024hydra} introduces sequential dependence between these heads, and D-Flash \citep{chen2026dflash} uses diffusion models to predict sequences in one forward pass. We focus on EAGLE-3 as the dominant auto-regressive drafter design.

\textbf{Attention sinks and patterns.} \citet{streamingllm} was the first to observe attention sinks and show that preserving these sinks is essential for long-context perplexity under sliding-window attention. \citet{gatedattention} proposed gated attention as a mechanism to reduce the model's reliance on a fixed sink token. Both works study sink behavior in the target model itself; we are the first, to our knowledge, to characterize attention behavior in speculative decoding drafters and identify a distinct failure mode (attention drift) specific to the chain-residual structure of these models.

\textbf{Normalization in transformers.} The placement of normalization relative to residual connections has been studied extensively in the standard transformer setting. \citet{xiong2020layer, ontheroleofattention} showed that pre-norm enables more stable training than post-norm, leading most modern LLMs to adopt pre-norm. Our work revisits this question for speculative decoding drafters, where iterative self-feedback across speculation steps changes the dynamics. We show that post-norm becomes necessary to prevent magnitude accumulation that drives attention drift.

% where the iterative self-feedback structure means standard pre-norm wisdom does not directly apply: we show that post-norm on the chain residual is necessary to prevent the magnitude accumulation that drives attention drift.

% We focus on EAGLE-3 as the dominant autoregressive drafter design in current production systems such as vllm, sglang; the chain-residual structure we study is shared by other autoregressive drafter variants. %, so the mechanism we identify likely generalizes beyond EAGLE-3.

\section{Limitations}

Our study has two main limitations. First, we focus on EAGLE-3 as the dominant post-training auto-regressive drafter in production. We observe attention drift in MTP heads, suggesting the phenomenon extends beyond EAGLE-3, but we do not investigate the alternative mechanisms at play there or evaluate our fix across other drafter designs. Extending the analysis to MTP, Hydra, and other variants is a natural direction for future work. Second, our experiments are limited to models below 120B parameters due to compute constraints; behavior at larger scale remains unverified.

\section{Conclusion}

We identified \textit{attention drift} in auto-regressive speculative decoding drafters: as the drafter generates successive tokens during speculation, attention progressively migrates from the sink onto recently-generated tokens. We characterized this phenomenon across model families and showed that it is closely tied to monotonic magnitude growth in the unnormalized residual path between speculation steps, which causes the drafter to behave more like additional transformer layers stacked on the target than as a standalone auto-regressive predictor. Two simple architectural interventions, post-norm on the drafter output and RMSNorm before the target hidden projection, address the underlying dynamics and yield consistent improvements, with the largest gains under deployment shift. 
Crucially, our experiments show that fixing drift alone is not enough to improve performance; \textbf{attention drift is merely a symptom of a deeper issue, not its cause}. Post-norm helps because it addresses the underlying dynamics: it prevents magnitude accumulation and stops the drafter from learning depth-specific transformations, which together improve both performance and robustness.
We hope that naming attention drift gives the community a useful diagnostic for drafter robustness, and that post-norm proves a useful default for production systems.

\clearpage

\begin{ack}
We thank fal and Lambda for the compute grants that supported this research.
\end{ack}

\bibliographystyle{unsrtnat}
\bibliography{references}

%%%%%%%%%%%%%%%%%%%%%%%%%%%%%%%%%%%%%%%%%%%%%%%%%%%%%%%%%%%%

\appendix

\newpage

% \section{Attention Drift}
% \label{ap:attention_drift}

% In this appendix, we provide additional attention-drift visualizations and analyze whether the phenomenon appears beyond EAGLE3 drafters.

%Another speculative argument for why attention drift happens concerns how speculative decoding models are trained. Most EAGLE-trainers during TTT use a fixed window size and repeatedly drops tokens from the left during generation. This could be a factor that makes the sink tend to weaken with the increasing depth as it gets pushed out.

% \subsection*{MTP Heads}
% \label{ap:mtp}

\section{Gated attention}
\label{ap:gated}

In the gated attention variant of our models, we add an optional per-element sigmoid gate over the hidden dimension to the EAGLE3 draft self-attention. Let $\mathbf{x} \in \mathbb{R}^{T \times d_{\mathrm{in}}}$ denote the input stream to the attention block. Because EAGLE3 fuses the verifier hidden state with the input-embedding inside the decoder layer, $\mathbf{x}$ is the concatenation of the two and $d_{\mathrm{in}} = 2d$, where $d$ is the draft hidden size. The query, key and value projections operate on $\mathbf{x}$ as in standard Llama, and we write the multi-head attention output \emph{before} the output projection as

\[
    \mathbf{A} \;=\; \mathrm{Attn}(\mathbf{x}\mathbf{W}_Q,\mathbf{x}\mathbf{W}_K,\mathbf{x}\mathbf{W}_V)
    \;\in\; \mathbb{R}^{T \times h\,d_h},
\]

with $h$ heads of dimension $d_h$. Our modification introduces a learned gate $\mathbf{W}_g \in \mathbb{R}^{d_{\mathrm{in}} \times h\,d_h}$ that shares its input with the QKV projections,

\begin{equation}
    \mathbf{g} \;=\; \sigma\!\big(\mathbf{x}\mathbf{W}_g\big),
\end{equation}

where $\sigma(\cdot)$ denotes the element-wise sigmoid ($\sigma(z) = \dfrac{1}{1 + e^{-z}}$). The gate is applied to the concatenated head output and \emph{precedes} the output projection $\mathbf{W}_O$:

\begin{equation}
    \mathbf{y} \;=\; \big(\mathbf{g} \odot \mathbf{A}\big)\,\mathbf{W}_O.
    \label{eq:gated_attn}
\end{equation}

\section{Benchmark Results}
\label{ap:results}

We report SGLang acceptance length (including the bonus token) per benchmark for each target model trained in this work. Tables 6–9 cover GPT-oss 20B, Qwen3.5 9B, Qwen3 8B, and Llama 3.1 8B respectively, sweeping the relevant decoding modes for each (low/medium reasoning effort for GPT-oss; thinking/no-thinking for Qwen3 and Qwen3.5). Across all four models and every decoding mode, post-norm matches or improves on pre-norm; the single regression is Llama 3.1 8B on MATH-500, which falls within evaluation noise. Gains are largest on math and code (e.g., GSM8K +10\% on GPT-oss low, HumanEval +5\% on Qwen3 8B no-think) and smallest on Alpaca-style open-ended chat, consistent with the harder, longer-horizon completions benefiting more from a stable drafter.

\begin{table}[h!]
\caption{SGLang acceptance length for \textbf{GPT-oss 20B} + EAGLE3 across pre/post-norm draft and low/medium reasoning effort. Best per row in bold.}
\centering
\small
\begin{tabular}{l r r r r}
\toprule
Benchmark & Pre, low & Post, low & Pre, medium & Post, medium \\
\midrule
MT-Bench & 2.59 & \textbf{2.65} & 2.52 & 2.64 \\
GSM8K & 3.68 & \textbf{4.04} & 3.54 & 3.73 \\
Alpaca & 2.49 & \textbf{2.55} & 2.48 & 2.55 \\
HumanEval & 3.62 & \textbf{3.80} & 3.08 & 3.27 \\
MATH-500 & 3.73 & \textbf{3.92} & 3.36 & 3.48 \\
GPQA & 2.57 & \textbf{2.73} & 2.45 & 2.58 \\
LiveCodeBench & 3.12 & \textbf{3.26} & 2.69 & 2.79 \\
\bottomrule
\end{tabular}
\label{tab:sglang_gptoss}
\end{table}
\begin{table}[h!]
\caption{SGLang acceptance length for \textbf{Qwen3.5 9B} + EAGLE3 across pre/post-norm draft and thinking/no-thinking decoding. Best per row in bold.}
\centering
\small
\begin{tabular}{l r r r r}
\toprule
Benchmark & Pre, no-think & Post, no-think & Pre, think & Post, think \\
\midrule
MT-Bench & 3.08 & 3.16 & 3.60 & \textbf{3.69} \\
GSM8K & 4.21 & 4.28 & 4.32 & \textbf{4.42} \\
Alpaca & 2.73 & 2.79 & 3.49 & \textbf{3.57} \\
HumanEval & 4.47 & \textbf{4.49} & 4.43 & 4.43 \\
MATH-500 & 5.08 & \textbf{5.12} & 4.95 & 5.01 \\
GPQA & 3.54 & 3.65 & 3.78 & \textbf{3.87} \\
LiveCodeBench & 4.31 & \textbf{4.36} & 4.22 & 4.30 \\
\bottomrule
\end{tabular}
\label{tab:sglang_qwen35}
\end{table}
\begin{table}[h!]
\caption{SGLang acceptance length for \textbf{Qwen3 8B} + EAGLE3 across pre/post-norm draft and thinking/no-thinking decoding. Best per row in bold.}
\centering
\small
\begin{tabular}{l r r r r}
\toprule
Benchmark & Pre, no-think & Post, no-think & Pre, think & Post, think \\
\midrule
MT-Bench & 3.55 & \textbf{3.61} & 3.09 & 3.15 \\
GSM8K & 5.58 & \textbf{5.73} & 3.73 & 3.83 \\
Alpaca & 3.14 & \textbf{3.35} & 2.68 & 2.82 \\
HumanEval & 5.00 & \textbf{5.21} & 3.05 & 3.13 \\
MATH-500 & 5.80 & \textbf{5.89} & 3.57 & 3.64 \\
GPQA & 3.91 & \textbf{4.04} & 2.98 & 3.08 \\
LiveCodeBench & 4.59 & \textbf{4.71} & 3.06 & 3.14 \\
\bottomrule
\end{tabular}
\label{tab:sglang_qwen3}
\end{table}
\begin{table}[h!]
\caption{SGLang acceptance length per benchmark for \textbf{Llama 3.1 8B} with EAGLE3 (steps=7, topk=1, draft=8). Best per row in bold. The single regression on MATH-500 is within evaluation noise.}
\centering
\small
\begin{tabular}{l r r}
\toprule
Benchmark & Pre-norm & Post-norm \\
\midrule
MT-Bench & 3.84 & \textbf{3.90} \\
GSM8K & 5.28 & \textbf{5.33} \\
Alpaca & 3.52 & \textbf{3.59} \\
HumanEval & 5.56 & \textbf{5.59} \\
MATH-500 & \textbf{5.87} & 5.81 \\
GPQA & 4.74 & \textbf{4.78} \\
LiveCodeBench & 4.16 & \textbf{4.22} \\
\bottomrule
\end{tabular}
\label{tab:sglang_llama}
\end{table}
\FloatBarrier

% \subsection{Model Template Sensitivity}
% \label{sec:templsens}

% Another factor in template sensitivity may lie in how the loss is constructed during training. In the EAGLE series, the loss is computed only on assistant tokens; user tokens don't contribute to the loss directly. So while parameters are shared and trained on the full sequence, the model's outputs at user positions are not directly constrained, only their utility for downstream assistant-position predictions is. This asymmetry could show up especially when the chat template is changed and the boundary between supervised and unsupervised positions change from where the model was trained to expect it.

% \input{tables/exp1_template_sensitivity_llama3_8b}
% \input{tables/exp1_template_sensitivity_qwen3_8b}
% \input{tables/exp1_template_sensitivity_gpt_oss_20b}
% \FloatBarrier

% \subsection{Model Prompt Length Sensitivity}

% \input{tables/exp2_prompt_length_llama3_8b}
% \input{tables/exp2_prompt_length_qwen3_8b}
% \input{tables/exp2_prompt_length_gpt_oss_20b}
% \FloatBarrier

\subsection{Model Long-Context SWA Sensitivity}

To our surprise, the Gated Post-Norm model did not suffer at all from out-of-distribution inference lengths and showed no accuracy loss in the full-context attention case. This could be attributed to its low attention entropy: it learns to attend to only a couple of tokens, and this sharpened attention pattern may filter the destructive signal via the sigmoid gate. This also suggests the model relies less on absolute positional information than the baseline does.

\begin{table}[h!]
\centering
\caption{\textbf{Llama 3.1 8B}: Avg accepted draft tokens per round at 8k context, comparing full attention to sliding-window variants with optional BoS / system-prompt carry.}
\label{tab:exp3}
\small
\begin{tabular}{l r r r r}
\toprule
Condition & pre-norm & post-norm & Gated pre-norm & Gated post-norm \\
\midrule
full attn & 0.05 $\pm$ 0.00 & 0.83 $\pm$ 0.04 & 0.84 $\pm$ 0.06 & \textbf{2.74} $\pm$ 0.17 \\
SWA & 2.43 $\pm$ 0.14 & \textbf{2.97} $\pm$ 0.16 & 1.21 $\pm$ 0.09 & 2.70 $\pm$ 0.17 \\
SWA + BoS & 2.53 $\pm$ 0.15 & \textbf{2.99} $\pm$ 0.16 & 1.22 $\pm$ 0.09 & 2.69 $\pm$ 0.17 \\
SWA + system & 0.06 $\pm$ 0.00 & \textbf{3.01} $\pm$ 0.15 & 1.72 $\pm$ 0.12 & 2.70 $\pm$ 0.17 \\
\bottomrule
\end{tabular}
\end{table}
\begin{table}[h!]
\centering
\caption{\textbf{GPT-oss 20b}: Avg accepted draft tokens per round at 16k context, comparing full attention to sliding-window variants.}
\label{tab:exp3}
\small
\begin{tabular}{l r r}
\toprule
Condition & Pre-norm & Post-norm \\
\midrule
full attn & \textbf{1.67} $\pm$ 0.09 & 1.67 $\pm$ 0.09 \\
SWA & 1.09 $\pm$ 0.08 & \textbf{1.94} $\pm$ 0.11 \\
\bottomrule
\end{tabular}
\end{table}
\begin{table}[h!]
\caption{\textbf{Qwen3-8B}: Avg accepted draft tokens per round at 8k context, comparing full attention to sliding-window variants with optional BoS / system-prompt carry.}
\centering
\small
\begin{tabular}{l r r}
\toprule
Condition & Pre-norm & Post-norm \\
\midrule
full attn & \textbf{1.33} $\pm$ 0.04 & 1.24 $\pm$ 0.04 \\
SWA & 1.87 $\pm$ 0.07 & \textbf{2.05} $\pm$ 0.07 \\
SWA + BoS & 1.97 $\pm$ 0.07 & \textbf{2.08} $\pm$ 0.07 \\
SWA + system & 1.87 $\pm$ 0.07 & \textbf{2.05} $\pm$ 0.07 \\
\bottomrule
\end{tabular}
\label{tab:exp3}
\end{table}
\FloatBarrier

\subsection{SWA Size / Accuracy}

Section 5.2 reports the window-size sweep for Llama 3.1 8B at 8k context. Table 20 confirms the same pattern holds for Qwen3 8B at a 32k target context: post-norm leads pre-norm at every window size, and gains saturate around w=1024 with only marginal improvement at w=2048.

\begin{table}[h!]
\centering
\caption{\textbf{Llama-3.1 8B}: Avg accepted draft tokens per round under SWA + BoS carry as the window size grows from 256 to 2048 (target context 8192).}
\label{tab:exp4}
\small
\begin{tabular}{l r r}
\toprule
Condition & Pre-norm & Post-norm \\
\midrule
w=256 & 1.90 $\pm$ 0.11 & \textbf{2.37} $\pm$ 0.13 \\
w=1024 & 2.53 $\pm$ 0.15 & \textbf{2.99} $\pm$ 0.16 \\
w=2048 & 2.61 $\pm$ 0.15 & \textbf{3.01} $\pm$ 0.15 \\
\bottomrule
\end{tabular}
\end{table}
\begin{table}[h!]
\centering
\caption{\textbf{Qwen3-8B}: Avg accepted draft tokens per round under SWA + BoS carry as the window size grows from 256 to 2048 (target context 32768).}
\label{tab:exp4}
\small
\begin{tabular}{l r r}
\toprule
Condition & Pre-norm & Post-norm \\
\midrule
w=256 & 1.64 $\pm$ 0.08 & \textbf{1.82} $\pm$ 0.08 \\
w=1024 & 1.97 $\pm$ 0.09 & \textbf{2.14} $\pm$ 0.09 \\
w=2048 & 2.00 $\pm$ 0.09 & \textbf{2.16} $\pm$ 0.09 \\
\bottomrule
\end{tabular}
\end{table}
\FloatBarrier

\subsection{LongBench Results}

We further break down LongBench-E results for Llama 3.1 8B by prompt-length bucket, using each architecture's preferred SWA mode (SWA+BoS for pre-norm, SWA+System Prompt for post-norm. Post-norm outperforms pre-norm in every bucket from 0–4k all the way to 32–36k. This indicates that the post-norm advantage seen on the multi-turn benchmark in Section 5.2 transfers to true long-context tasks rather than being specific to the multi-turn setting.

\begin{table}[h!]
\centering
\small
\caption{\textbf{Llama 3.1 8B}, \textbf{LongBench-E}: avg accepted draft tokens per round vs prompt length, averaged over all tasks. Each model uses its preferred SWA mode (pre-norm: SWA + BoS, post-norm: SWA + Sys Prompt). $n$ is the number of verification rounds in each bucket.}
\begin{tabular}{r | cr | cr}
\toprule
Ctx (tokens) & \multicolumn{2}{c}{Pre-norm (SWA + BoS)} & \multicolumn{2}{c}{Post-norm (SWA + Sys Prompt)} \\
 & Acceptance Length ($\tau$) & Number of samples &Acceptance Length ($\tau$) & $n$ \\
\midrule
0-4k & 1.60 & 7838 & \textbf{1.99} & 6802 \\
4-8k & 1.63 & 11146 & \textbf{2.07} & 9563 \\
8-12k & 1.30 & 9927 & \textbf{1.73} & 8354 \\
12-16k & 1.28 & 6242 & \textbf{1.60} & 5482 \\
16-20k & 1.82 & 3000 & \textbf{1.98} & 2839 \\
20-24k & 1.94 & 3855 & \textbf{1.96} & 3831 \\
24-28k & 1.77 & 2744 & \textbf{2.20} & 2374 \\
28-32k & 1.73 & 1560 & \textbf{2.25} & 1309 \\
32-36k & 1.21 & 542 & \textbf{1.80} & 428 \\
\bottomrule
\end{tabular}
\label{tab:exp5_per_ctx}
\end{table}

\FloatBarrier

\section{Training}
\label{ap:training}

We have used a modified instance of the \textit{SpecForge} repository to train our models. This repository implements the Train-Time-Test method for EAGLE-3 training and it is the framework used to train current state-of-the-art EAGLE-3 models.

Llama and Qwen3 variants are trained on the Open-PerfectBlend dataset with regenerated answers using target model. The dataset consists over 1.4M samples. We have run it for 2 epochs and $1.5 \times 10^{-4}$ LR, with effective batch size of 4. On average training took around 48 H200 hours for each model with 8-9B parameters.

Qwen3.5 and Gpt-oss variants were trained on Nemotron post training dataset with regenerated answers using target model. The dataset consist over 1.4M samples. The maximum sequence length was 8K tokens. We have run it for 1 epoch and $1.5 \times 10^{-4}$ LR. On average training took around 36 to 48 H200 hours based on model size (20B and 120B).

\section{Template Perturbations}
\label{ap:template}

\begin{verbatim}
--- regular ---
<|begin_of_text|><|start_header_id|>system<|end_header_id|>

You are a helpful assistant.<|eot_id|><|start_header_id|>user<|end_header_id|>

What is the capital of France?<|eot_id|><|start_header_id|>assistant<|end_header_id|>


--- no_bos ---
<|start_header_id|>system<|end_header_id|>

You are a helpful assistant.<|eot_id|>...

--- no_template ---
<|begin_of_text|>Question: What is the capital of France?
Answer:

--- no_bos_no_template ---
Question: What is the capital of France?
Answer:
\end{verbatim}

%%%%%%%%%%%%%%%%%%%%%%%%%%%%%%%%%%%%%%%%%%%%%%%%%%%%%%%%%%%%

\end{document}